%% file: root.tex
\documentclass[letterpaper, 10 pt, conference]{ieeeconf}  

\IEEEoverridecommandlockouts                              

\usepackage{graphics} 
\usepackage{graphicx}
\usepackage{booktabs}
\usepackage{xparse} 
\usepackage{color, colortbl}
\usepackage{amsmath,amssymb,amsfonts}
\usepackage{textcomp}
\usepackage{xcolor}
\usepackage{multirow}
\usepackage{multicol}
\usepackage{subcaption}
\usepackage{amsmath} 
\DeclareMathOperator*{\argmax}{argmax}
\usepackage{amssymb}  
\usepackage{algorithm}
\usepackage[noend]{algpseudocode}
\usepackage{lipsum}
\usepackage{gensymb}

\usepackage{amsthm}
\usepackage{float}
\let\labelindent\relax
\usepackage[inline]{enumitem}
\usepackage{hyperref}
\graphicspath{{./figs/}{./figs/real_ns/}{./figs/real_s}{./figs/sim_vis}}

\newcommand{\ignore}[1]{}

\title{\LARGE \bf
Attribute-based Object Grounding and Robot Grasp Detection with Spatial Reasoning
}

\author{Houjian Yu$^{1}$, Zheming Zhou$^{2}$, Min Sun$^{2}$$^,$$^{3}$, Omid  Ghasemalizadeh$^{2}$, Yuyin Sun$^{2}$, \\ Cheng-Hao Kuo$^{2}$,  Arnie Sen$^{2}$, and Changhyun Choi$^{1}$
\thanks{*This work was supported by Amazon Lab126.}
\thanks{$^{1}$ The authors are with the Department of Electrical and Computer Engineering, Univ. of Minnesota, Minneapolis, USA
        {\tt\small \{yu000487, cchoi\}@umn.edu}}
\thanks{$^{2}$ The authors are with Amazon Lab126, Sunnyvale, CA, USA
        {\tt\small \{zhemiz, minnsun, ghasemal, yuyinsun, chkuo, senarnie\}@amazon.com}}
\thanks{$^{3}$ The authors is with National Tsing Hua University, Taiwan
        {\tt\small \{ sunmin\}@ee.nthu.edu.tw}}
}

\input{_macros}
\begin{document}

\maketitle
\thispagestyle{empty}
\pagestyle{empty}

\input{00_abstract}


\input{01_introduction}

\input{02_related}

\input{03_methods}

\input{04_experiments}

\input{05_conclusion}




{
\bibliography{IEEEexample,IEEEabrv}
\bibliographystyle{IEEEtran}
}

\end{document}

%% file: 00_abstract.tex
\begin{abstract} 

Enabling robots to grasp objects specified through natural language is essential for effective human–robot interaction, yet it remains a significant challenge. Existing approaches often struggle with open–form language expressions and typically assume unambiguous target objects without duplicates. Moreover, they frequently rely on costly, dense pixel–wise annotations for both object grounding and grasp configuration. We present Attribute–based Object Grounding and Robotic Grasping (OGRG), a novel framework that interprets open–form language expressions and performs spatial reasoning to ground target objects and predict planar grasp poses, even in scenes containing duplicated object instances. We investigate OGRG in two settings: (1) Referring Grasp Synthesis (RGS) under pixel–wise full supervision, and (2) Referring Grasp Affordance (RGA) using weakly supervised learning with only single–pixel grasp annotations. Key contributions include a bi-directional vision–language fusion module and the integration of depth information to enhance geometric reasoning, improving both grounding and grasping performance. Experiment results show that OGRG outperforms strong baselines in tabletop scenes with diverse spatial language instructions. In RGS, it operates at 17.59 FPS on a single NVIDIA RTX 2080 Ti GPU, enabling potential use in closed–loop or multi–object sequential grasping, while delivering superior grounding and grasp prediction accuracy compared to all the baselines considered. Under the weakly supervised RGA setting, OGRG also surpasses baseline grasp–success rates in both simulation and real–robot trials, underscoring the effectiveness of its spatial reasoning design. Project page: \url{https://z.umn.edu/ogrg}

\end{abstract}

%% file: 01_introduction.tex
\section{Introduction}
Target-oriented robot grasping is a fundamental task in robot manipulation, with wide-ranging applications in real-world scenarios. Compared to purely vision-driven robot grasping approaches~\cite{lou2023adversarial,yu2022self,yu2023iosg,kumra2020antipodal}, language-driven robot grasping offers greater flexibility by leveraging object attributes (e.g., color, shape, category name, and spatial location) to differentiate the target object from others~\cite{vuong2024language,yang2024attribute,xu2023joint}. This capability reduces ambiguity in target identification. However, the reliance on detailed language descriptions introduces additional challenges for robust vision–language understanding.

\begin{figure}[t]
\centering
\includegraphics[width=\linewidth]{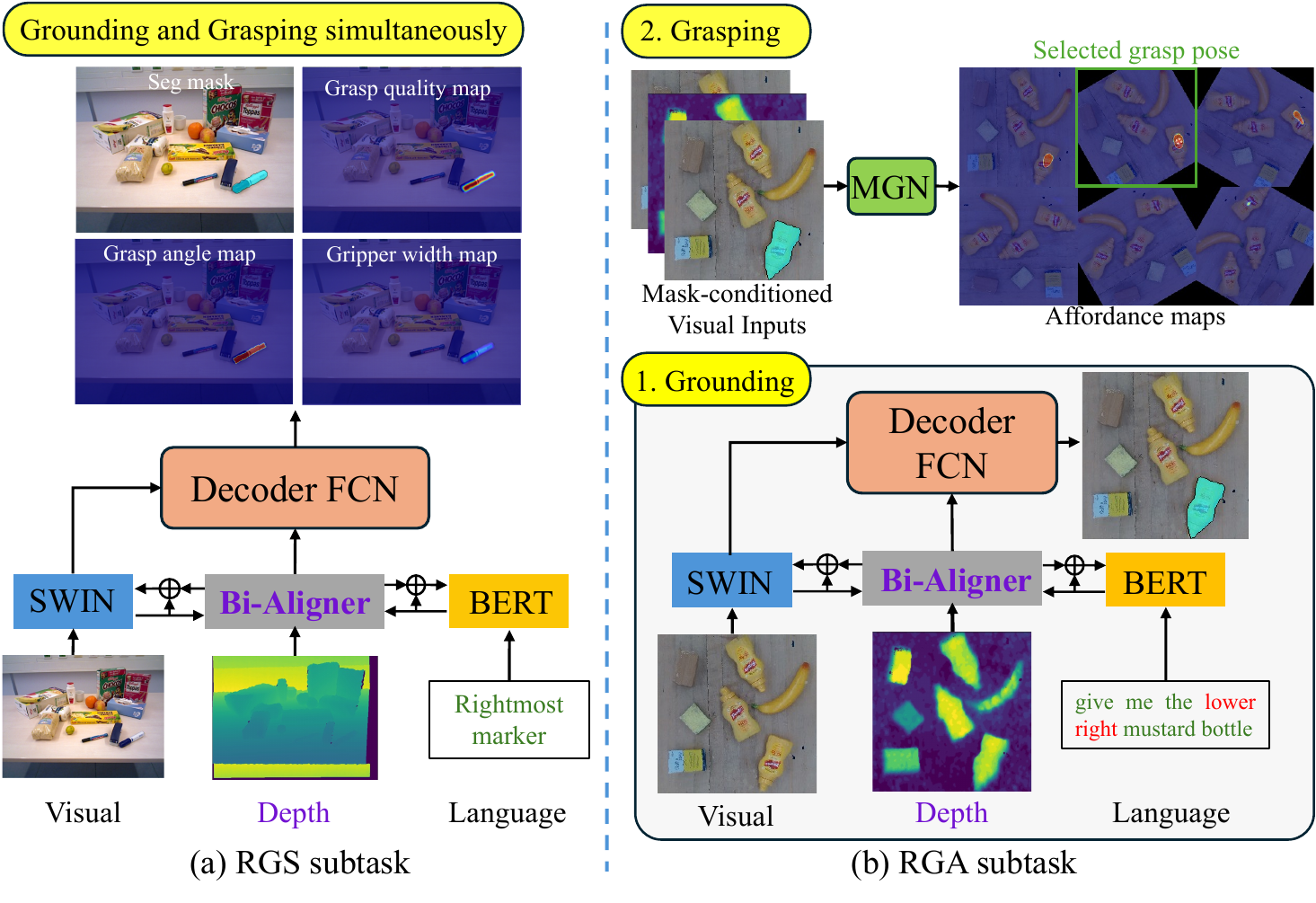}
\vspace{-5mm}
\caption{\textbf{Object Grounding and Robot Grasping (OGRG) model with open-form expressions for spatial reasoning.} The model is designed to solve the attribute-based grounding and grasp detection task. The RGS subtask aims at predicting grasp rectangles with pixel-wise full supervision. The RGA subtask focuses on predicting grasp affordances with weak grasping supervision.}
\label{fig:front}
\vspace{-5mm}
\end{figure}

Previous language-conditioned robot grasping approaches have been constrained to predefined vocabulary and simplistic attribute descriptions (e.g., ``red apple''), while typically assuming the presence of distinct, non-duplicated objects in the scene~\cite{yang2024attribute,xu2023joint,ahn2022visually}. Consequently, these methods cannot handle open-form language inputs or perform more challenging grounding tasks such as spatial reasoning. While recent advances in Multimodal Large Language Models (MLLMs), including Vision–Language–Model (VLM)  and Vision–Language–Action (VLA) models, have demonstrated strong multimodal understanding for high-level planning and action generation~\cite{jin2024reasoning,kim2024openvla,ma2024glover,zawalski2024robotic,tziafas2024openworldgraspinglargevisionlanguage,qian2024thinkgrasp}, their substantial computational demands for data generation, training, and inference often limit their deployment on resource-constrained robotic platforms. In addition, annotating large-scale robotic grasp datasets with detailed language descriptions remains labor-intensive for both human operators and autonomous systems.

Despite the promising performance with MLLMs across various objects and environments, two critical challenges remain:
\begin{itemize}
    \item \textbf{Grasp learning perspective}: \textit{Can a compact and computationally efficient fusion module, serving as an alternative to MLLMs, be designed to effectively align vision and language features for real-world robot grasping?}
    \item \textbf{Data efficiency perspective}: \textit{Can the grasping model be trained in a weakly supervised manner using sparse and imperfect labels?}
\end{itemize}

Without relying on pre-aligned vision–language models~\cite{radford2021learning,li2022blip}, we primarily investigate effective multimodal fusion and training with both dense and sparse labels for object grounding and grasp detection. In this paper, we propose a bi-directional multimodal fusion module to align vision, language, and depth features from different embedding spaces. The fused multimodal features are used for two grasp detection settings as shown in Fig .\ref{fig:front}: Referring Grasp Synthesis (RGS) \cite{yu2024parameter, tziafas2023language, vuong2024language} and Referring Grasp Affordance (RGA) \cite{yang2024attribute, yu2024parameter, mees2023grounding} (see \ref{prob_form} for problem formulation details). Both tasks require pixel-level vision-language understanding for object grounding and planar grasp pose prediction. The most closely related work to ours is ETRG \cite{yu2024parameter}, which employs the CLIP model with a downsampling-then-upsampling strategy for parameter-efficient tuning, aiming to reduce the number of trainable parameters while maintaining multimodal alignment. However, the aggressive feature downsampling inevitably leads to information loss, resulting in suboptimal performance on object grounding. In contrast, our approach strikes a better balance between model compactness (approximately 240M total parameters) and task performance by introducing a novel fusion module that facilitates effective interaction between the vision and language backbones.

Our grounding and grasping system is capable of (1) accepting open-form object attribute descriptions, including colors, shapes, category names, and spatial reasoning, to predict target object masks and planar grasp poses in the format of grasp rectangles and grasp affordance maps, (2) effectively fusing multimodal features without relying on pre-aligned models and utilizing both dense and sparse robot grasping labels for predictions, and (3) grounding the target object with high accuracy while achieving a high success rate in grasping.

Our primary contributions are summarized as follows:
\begin{itemize}
    \item An end-to-end Object Grounding and Robot Grasping (OGRG) model for RGS and RGA tasks, predicting object masks and 5-DoF grasp poses under dense supervision, and grasp affordances under weak supervision.
    \item A bi-directional multimodal fusion module is introduced to align vision, language, and OGRG downstream tasks, RGS and RGA.
    \item The approach is validated through comprehensive experiments in both simulation and real robot environments, demonstrating effectiveness across diverse objects and open-form language inputs compared to baseline methods.
\end{itemize}

%% file: 02_related.tex
\section{Related Work}

\subsection{Language-guided Object Grounding}
Language-guided object grounding focuses on localizing the target object referred to by language within an image. This problem can be categorized into two tasks based on the output type for localization: referring expression comprehension (REC), which predicts bounding boxes, and referring expression segmentation (RES), which generates binary segmentation masks. This paper concentrates on pixel-level multimodal alignment for predicting segmentation masks and grasping affordances, deriving from the RES task.

Early RES methods \cite{hu2016segmentation, li2018referring, liu2017recurrent} employed fully convolutional networks for visual feature extraction and RNN/LSTM architectures for language embeddings. These approaches typically utilized simple multimodal feature concatenation or multiplication, followed by convolutional layers and upsampling, to decode target masks. More recent works \cite{yu2024parameter, xu2023bridging, wangbarleria, wang2022cris} leverage well-aligned vision-language models, such as the CLIP model \cite{radford2021learning}, pretrained on large-scale datasets, to enhance vision-language fusion. Another line of research, including LAVT \cite{yang2022lavt}, CGFormer \cite{tang2023contrastive}, and DMMI \cite{hu2023beyond}, incorporates Swin Transformer \cite{liu2021swin} as the visual feature extractor while actively exploring cross-modal attention mechanisms to embed vision-language features into a shared space. The method LAVT \cite{yang2022lavt} is closely related to ours, which introduces early-stage uni-directional fusion modules for visual and linguistic feature interactions. In contrast, this work proposes bi-directional multimodal fusion modules that further integrate depth features, extending the RES task to the robotics domain for object grounding and grasp pose prediction.


\subsection{Language-guided Robot Grasping}
Recent advancements in vision-language models~\cite{radford2021learning, liu2024visual, touvron2023llama} have significantly advanced the field of language-guided robot grasping, enabling robots to identify and manipulate objects based on user-provided natural language instructions~\cite{xu2023joint,yu2024parameter,tziafas2023language, jin2024reasoning, vuong2024language,  mees2023grounding, shridhar2022cliport, kim2024openvla}. These models leverage pre-trained embeddings, such as CLIP and similar architectures, to align visual and textual modalities effectively, allowing robots to process diverse and open-vocabulary instructions. Early approaches primarily focused on tasks like object detection and segmentation, often relying on handcrafted features and task-specific training data.
Recent methodologies extend these capabilities by integrating multimodal transformers and attention mechanisms to enhance contextual understanding and reasoning~\cite{jin2024reasoning, vuong2024language}. These models excel at handling ambiguous or complex instructions, such as spatial references or multi-object contexts, by generating grasp affordance maps and candidate grasp poses with high precision. Such advancements pave the way for more robust and flexible applications in unstructured and dynamic environments, addressing key challenges in open-world robot manipulation.
Our work builds on previous vision and language models while targeting object grounding and robot grasping tasks with strong and weak supervision.

%% file: 03_methods.tex
\section{Method}
In this section, we propose the OGRG model for attribute-based language-guided object grounding and robot grasping. The OGRG method is designed to address two primary grasp detection tasks: Referring Grasp Synthesis (RGS) and Referring Grasp Affordance (RGA).

\begin{figure*}[!t]
\begin{center}
    \includegraphics[width=0.61\linewidth]{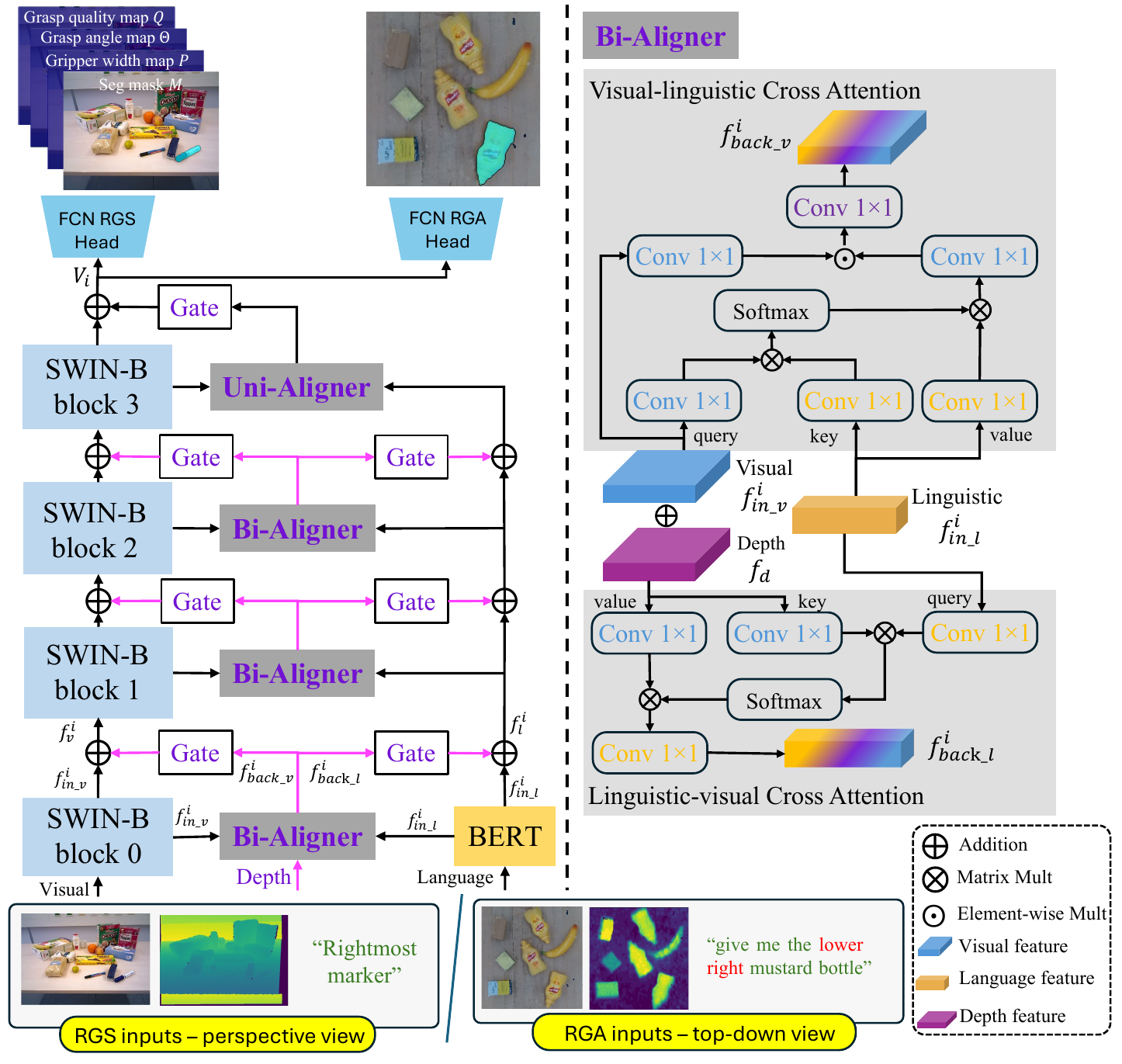}
\end{center}
\vspace{-2mm}
\caption{\textbf{OGRG Architecture.} OGRG processes open-form language expressions, visual images, and depth maps as inputs to generate task predictions. The bidirectional aligner (Bi-Aligner) fuses the multimodal features extracted from Swin Transformer \cite{liu2021swin} at different stages and the BERT language model \cite{devlin2018bert}. The updated multimodal features $f_{back\_v}^i$ and $f_{back\_l}^i$ are fed back into their corresponding visual and linguistic branches via feature gates. Finally, the light-weighted fully convolutional network (FCN) heads processes the updated visual features at different stages to produce the task-specific outputs.}
\label{fig:sys}
\vspace{-4mm}
\end{figure*}

\subsection{Problem Formulation} \label{prob_form}
The detailed formulations for the RGS and RGA subtasks are first introduced, as shown in Fig. \ref{fig:front}. These tasks involve processing multimodal inputs, including an RGB image denoted as $I \in \mathbb{R}^{H\times W\times 3}$, where $(H, W)$ represent the image dimensions; an attribute-based language description with token length $L$ represented as $T \in \mathbb{R}^L$; and a depth image denoted as $D \in \mathbb{R}^{H\times W}$. The proposed OGRG model leverages these visual observations and open-form language expressions to generate predictions for object grounding binary masks $M \in \mathbb{R}^{H\times W}$ and robot grasp pose $G$. We use a unified representation for grasp detection here: $G = \{x, y, z, \theta, l\}$ is parametrized by: $(x, y)$, the gripper center location in the image coordinate; $z$, the depth value; $\theta$, the gripper rotation angle in the camera frame; and $l$, the open width of the gripper. 

\textbf{RGS Subtask:} This task aims to predict a segmentation mask and planar 5-DoF grasp poses in the format of grasp rectangles, given an RGB image $I$, a depth map $D$ from the camera perspective view, and a language expression $T$. Following \cite{tziafas2023language}, the OGRG model predicts three grasp-related maps to reconstruct the gripper pose. The ground-truth grounding and grasping maps are from the OCID-VLG dataset \cite{tziafas2023language}. The system is formulated as $\Phi(I, D, T) = \{M, Q, \Theta, P\}$ to predict the object grounding mask $M$, the grasp quality map $Q$, the grasp angle map $\Theta$, and the gripper open width map $P$. Specifically, $(x, y)$ is determined by the pixel coordinate of the maximum value in $Q$. The rotation angle $\theta$ is derived as $\theta = \Theta(x, y)$ from $\Theta$. $z$ can be derived from depth map $D(x,y)$. Finally, the gripper open width $l$ is obtained from $P$ as $l = P(x, y)$.

\textbf{RGA Subtask:} Unlike RGS, RGA will predict grasp affordance maps $A \in \mathbb{R}^{H \times W \times N}$ with $N$ discretized rotation angles~\cite{yang2024attribute} (Fig. \ref{fig:front}b). We adopt a segment-then-grasp pipeline with OGRG and a Mask-Conditioned Grasping Network (see section \ref{MGN section}). The 5-DoF grasp pose is derived from the affordance maps, where $(x^*, y^*, \theta^*) = \argmax_{(x, y, \theta)} A(x, y, \theta)$ and $z = D(x^*, y^*)$. Here, $(x^*, y^*)$ corresponds to the pixel coordinate with the maximum affordance value, $\theta^*$ represents the optimal rotation angle, and $z$ provides the depth value for the grasp motion. We use a predefined gripper open width $l^*$ for all grasp attempts in RGA to facilitate the data collection process. 

\subsection{OGRG Multimodal Feature Fusion}

Fig. \ref{fig:sys} illustrates the details of the proposed OGRG model. Depending on the different settings for RGS and RGA subtasks, the OGRG model will provide 4 different grounding and grasping maps $\{M, Q, \Theta, P\}$ simultaneously for the RGS subtask after passing the task-specific FCN head. On the other hand, the OGRG model will only predict the object grounding map for RGA subtask.

The OGRG model employs Swin Transformer \cite{liu2021swin} as the visual backbone and BERT Transformer \cite{devlin2018bert} as the language feature extractor. For the depth branch, a ResNet-18 \cite{he2016deep} model is utilized to extract depth features $f_d$. To enable efficient vision-language alignment, the model incorporates a four-stage hierarchical multimodal fusion process with multiple aligners. At each stage, the visual features $f_{in\_v}^i \in \mathbb{R}^{C_i \times H_i \times W_i}$ and linguistic features $f_{in\_l}^i \in \mathbb{R}^{L \times C_t}$ interact through two distinct cross-attention mechanisms, resulting in fused multimodal features $f_{back\_v}^i$ and $f_{back\_l}^i$ for their respective branches. Here, $C_i$, $H_i$, and $W_i$ represent the number of channels, height, and width of the $i$-th stage ($i \in \{1, 2, 3, 4\}$), while $L$ and $C_t$ denote the language token length and token dimension, respectively.

The fused features are passed through learnable feature gates $g_i$ and added element-wise to $f_{in\_v}^i$ and $f_{in\_l}^i$, generating enhanced visual and linguistic features $f_v^i$ and $f_l^i$. Finally, the light-weighted Fully Convolutional Network (FCN) head processes the four-stage intermediate multimodal visual feature maps to produce the final task-specific outputs.

\subsection{Bidirectional Aligner} 
Inspired by the unidirectional fusion module from LAVT \cite{yang2022lavt}, a bidirectional aligner is proposed to update the two branches simultaneously. The bidirectional aligner (Bi-Aligner in Fig. \ref{fig:sys}) consists of visual-linguistic and linguistic-visual cross-attention mechanisms for multimodal fusion. The visual and depth features are first fused via element-wise addition, $f_{in\_vd}=f_{in\_v} + f_d$, where the depth feature $f_d$ is used only at the first stage. Given the flattened visual-depth features $f_{in\_vd}^i \in \mathbb{R}^{C_i \times D}$, where $D=H_i\times W_i$, and the linguistic features $f_{in\_l}^i \in \mathbb{R}^{L \times C_t}$ from the model backbone, cross-attention features are computed using the transformer attention formulation:
\begin{align}
    f_{cross\_v}^i & = \text{softmax}\left(\frac{(W_q^{V} f_{in\_vd}^i)^{T} (W_k^{V} f_{in\_l}^i)}{\sqrt{C_i}}\right)(W_v^{V} f_{in\_l}^i)^{T}, \\
    f_{cross\_l}^i & = \text{softmax}\left(\frac{(W_q^L f_{in\_l}^i)^{T} (W_k^L f_{in\_vd}^i)}{\sqrt{C_t}}\right)(W_v^L f_{in\_vd}^i)^{T},
\end{align}
where $W_q^{V}, W_k^{V}, W_v^{V}, W_q^{L}, W_k^{L}, W_v^{L}$ are projection matrices that unify the visual-depth and linguistic feature dimensions. The resulting cross-modal features are reshaped into $f_{cross\_v}^i \in \mathbb{R}^{C_i \times H_i \times W_i}$ and $f_{cross\_l}^i \in \mathbb{R}^{L \times C_t}$. These features are further processed with $1\times 1$ convolutions and ReLU activations to produce the fused visual and linguistic features, $f_{back\_v}^i$ and $f_{back\_l}^i$. 

Following the language pathway design from LAVT \cite{yang2022lavt}, learnable gates $g_i$ are applied to enhance the features, yielding the final visual and linguistic outputs:
\begin{align}
    f_v^i & = f_{in\_v}^i + g_i(f_{back\_v}^i), \\
    f_l^i & = f_{in\_l}^i + g_i(f_{back\_l}^i).
\end{align}

\subsection{Task Specific FCN Head}
The annotation $V_i$, $i \in \{1, 2, 3, 4\}$ is used to represent the intermediate visual features as inputs to the FCN head. From an empirical result, we use $V_i = \{f_{v}^i\}$ in RGS, and $V_i = \{f_{back\_v}^i\}$ in RGA for best performance. The decoding process is formulated as:
\begin{align}
\left\{ 
\begin{array}{ll}
Y_4 &= V_4, \\
Y_i &= \text{Conv}([\text{Up}(Y_{i+1}); V_i]), \quad i = 3, \; 2, \; 1,
\end{array}
\right.
\end{align}
where $\text{Conv}(\cdot)$ denotes a $3\times 3$ convolution layer followed by batch normalization and a ReLU activation function, and $\text{Up}(\cdot)$ indicates bilinear interpolation upsampling. The decoded $Y_1$ serves as the final model prediction, outputting $\{M, Q, \Theta, P\}$ for RGS (conducting grounding and grasping simultaneously) and $M$ for RGA (only for object grounding). 

To train the OGRG-RGS model, the loss function combines the cross-entropy loss for object grounding mask prediction ($M$) with smooth L1 losses for the grasp quality map ($Q$), grasp angle map ($\Theta$), and gripper open width map ($P$). For the OGRG-RGA model, dice loss and focal loss are applied for object grounding mask prediction.

\subsection{Mask-conditioned Grasping Network for RGA} \label{MGN section}
\begin{figure}[t]
\centering
\includegraphics[width=\linewidth]{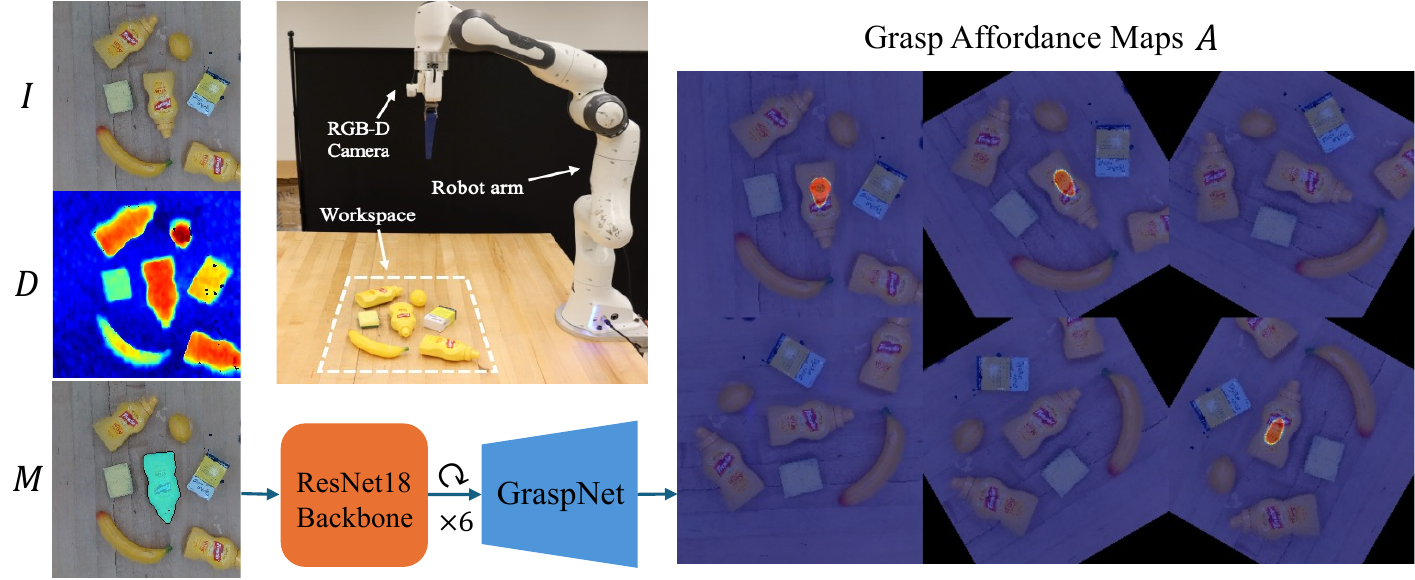}
\caption{\textbf{Mask-conditioned Grasping Network (MGN).} Conditioned on the object grounding mask $M$ predicted from OGRG, the MGN network uses a fully convolutional encoder-decoder architecture for pixel-level grasp affordance prediction with different rotation angles.}
\label{fig:mgn}
\vspace{-4mm}
\end{figure}

As shown in Fig. \ref{fig:mgn}, Mask-conditioned Grasping Network (MGN) predicts pixel-level affordance maps $A$ for $N = 6$ discrete rotation angles, each a multiple of $\theta = 30^\degree$ based on the RGB image $I$, the depth map $D$, and the OGRG grounding mask $M$. Specifically, the inputs are concatenated and passed through a ResNet-18 backbone. To handle challenging grasping rotations, tensor transformations are applied, and the processed features are fed into an FCN-based GraspNet, which consists of standard convolutional layers, batch normalization, and ReLU activations. A Sigmoid layer serves as the final output layer to produce the grasp affordances.

To minimize human annotation effort, this problem is formulated in a weak-supervision manner, where only a binary $\{0, 1\}$ ground-truth label is provided for the sampled grasp location—a single pixel among the $N$ rotation maps—while other pixels remain unlabeled. The training process employs a motion loss $\mathcal{L}_{grasp}$ from Attribute-Grasp \cite{yang2024attribute}

\subsection{Dataset Collection for RGA} \label{MGN section}
We collect training and testing data for the RGA task in the CoppeliaSim simulator~\cite{6696520}. From a pool of 32 objects, 7 were randomly selected—primarily from the YCB dataset~\cite{calli2015ycb}—to construct grasping scenes, as illustrated in Fig.~\ref{fig:testing_sim}. To improve robustness to environmental variations, we applied domain randomization to the background textures. Language instructions were generated from multiple templates incorporating object color, shape, category name, and spatial location. Spatial relationships were expressed in two forms: \textit{Absolute} (relative to the workspace) and \textit{Relative} (relative to other reference objects), increasing the variety of spatial reasoning cases. In total, we collected over $16{,}000$ visual–language–grasp triplets for model training.

%% file: 04_experiments.tex
\section{Experiments}
%
In this section, attribute-based object grounding and robot grasping experiments are conducted to evaluate the proposed method. The objectives of the experiments are to verify the following:  
1) The Bi-Aligner with depth fusion effectively fuses multimodal features without relying on pre-aligned vision-language backbones.  
2) The OGRG model demonstrates the ability to understand object-attribute descriptions and achieves a high grasping success rate, even with complex spatial relationship language inputs.  
3) The proposed OGRG-RGA, combined with the MGN, successfully addresses the weakly supervised RGA problem and efficiently adapts from simulation to real robot experiments.

\subsection{OGRG-RGS with Depth Fusion}
\noindent\textbf{Implementation Details:} The OGRG-RGS model is trained on the OCID-VLG dataset for 26 epochs with a batch size of 4 per GPU, using a total of 8 NVIDIA V100 GPUs. The training process employs the AdamW optimizer with an initial learning rate of $\lambda=0.00005$ and a polynomial learning rate decay. For fair comparisons, input images are resized to a resolution of $416 \times 416$, and the maximum sentence length is capped at 20 tokens for all baselines.

\noindent\textbf{RGS Evaluation Dataset:} The OGRG-RGS model and corresponding baselines are evaluated on the OCID-VLG dataset \cite{tziafas2023language}. This dataset is designed for target object grounding and grasp pose prediction based on open-form language descriptions. It includes 58 unique object candidates, over $89.6k$ referring language expressions describing a wide range of object attributes, and more than $75k$ hand-annotated grasp rectangles.

\noindent\textbf{Evaluation Metrics:} The image segmentation results for the language-referred object grounding task are evaluated using the mean intersection over union (mIoU). For robot grasping, the Jaccard Index $J@N$ metric, as described in \cite{tziafas2023language}, is employed. This metric measures the top-$N$ grasp rectangles that achieve an IoU greater than 0.25 and have rotation angle differences of less than $30^{\circ}$ compared to the ground-truth grasp rectangles.

\noindent\textbf{Baselines:} The RGS evaluation results are reported with the following baselines:  
1) CROG, proposed by \cite{tziafas2023language}, extends the referring expression segmentation model CRIS \cite{wang2022cris} for grasp map prediction. This approach involves full model fine-tuning, including the pre-trained CLIP model \cite{radford2021learning}.  
2) ETRG \cite{yu2024parameter} is a CLIP-based method that employs a parameter-efficient tuning framework with depth fusion branches. Instead of fine-tuning the full CLIP model, it uses a bidirectional adapter optimized for multiple tasks.  
3) HiFi-CS \cite{lu2023vl} applies hierarchical FiLM \cite{perez2018film} fusion for multimodal alignment and serves as another CLIP-based object grounding method.
4) LAVT \cite{yang2022lavt} adopts unidirectional aligners for multimodal fusion.
5) OGRG-nodepth applies our bi-directional aligner but without depth inputs.

\begin{table}[t]
    \centering
    \caption{\textbf{OGRG-RGS ablation study and baseline comparison on the OCID-VLG dataset.} Our proposed method improves both object grounding performance and the accuracy of grasp rectangle predictions.}
    \begin{tabular}{l|c|c|c}
        \hline
        \multirow{2}{*}{Baselines} & \multicolumn{1}{c|}{Grounding} & \multicolumn{2}{c}{Grasping} 
        \\
        \cline{2-4}
        & mIoU & J@1 & J@Any \\
        \hline
        CROG \cite{tziafas2023language}  & 81.10 & 77.20 & 87.70 \\
        ETRG \cite{yu2024parameter}  & 80.11 & 89.38 & 93.49 \\
        HiFi-CS \cite{lu2023vl} & 88.26 & - & - \\ 
        LAVT \cite{yang2022lavt}  & 92.52 & 87.55 & 91.77 \\
        OGRG-nodepth  & 94.87 & 88.49 & 93.70 \\
        \textbf{OGRG (Ours)}  & \textbf{95.60} & \textbf{90.81} & \textbf{94.70} \\
        \hline
    \end{tabular}
    \vspace{-4mm}
    \label{tab:rgs_comparison}
\end{table}

\noindent\textbf{RGS Results.} Table \ref{tab:rgs_comparison} presents the language-guided object grounding and robot grasping performance comparison with the selected baselines on the OCID-VLG dataset \cite{tziafas2023language}. The proposed OGRG-RGS consistently outperforms all baselines across different backbone architectures. For the object grounding task, the method achieves a significant improvement of $+14.5\%$ mIoU compared to CROG. Additionally, the Bi-Aligner with depth fusion enhances grounding performance by $+3.08\%$ and $+0.73\%$ mIoU compared to LAVT and OGRG-nodepth. In terms of robot grasping performance, OGRG-RGS demonstrates a substantial improvement, achieving $+13.61\%$ $J@1$ over CROG. The ablation study further highlights that both the Bi-Aligner and depth fusion contribute significantly to the overall accuracy of grasp rectangle predictions. During model inference, OGRG is able to run on a single RTX 2080Ti GPU with an inference speed of 17.59 FPS.

\subsection{Referring Grasp Affordance with Weak Supervision}

\begin{figure}[t]
  \centering
  \begin{subfigure}{0.28\textwidth}
    \includegraphics[width=\textwidth]{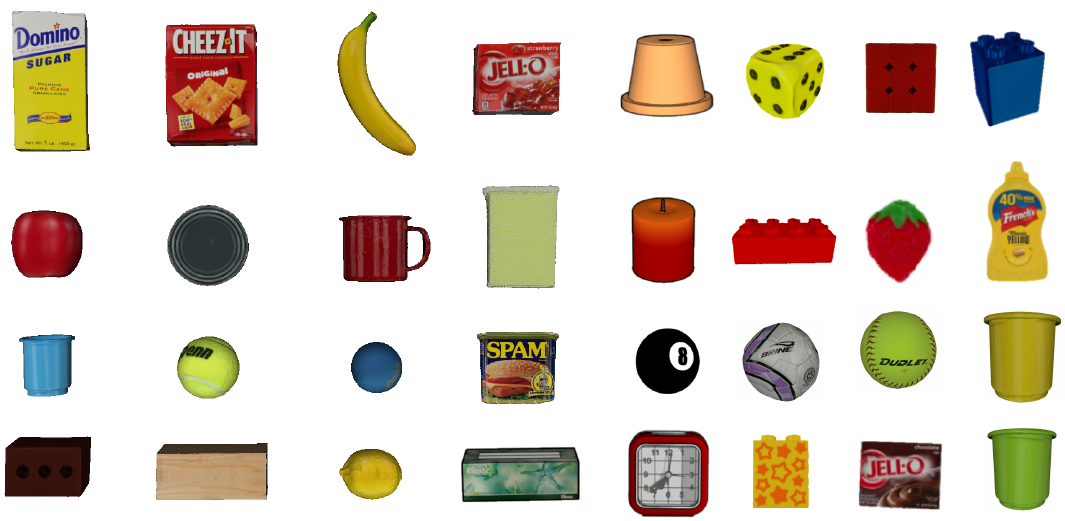}
    \caption{Simulated novel objects}
    \label{fig:testing_sim}
  \end{subfigure}
  \hfill
  \begin{subfigure}{0.19\textwidth}
    \includegraphics[width=\textwidth]{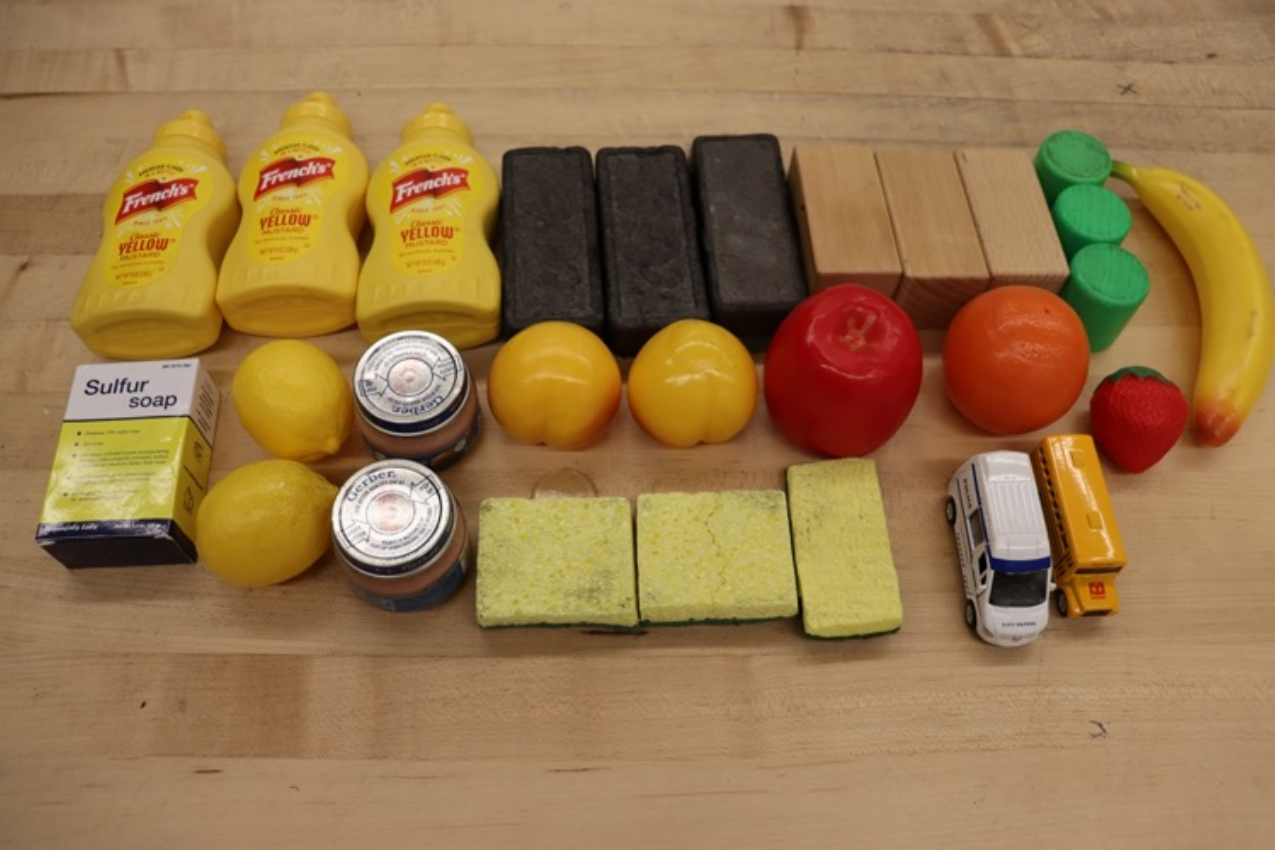}
    \caption{Real-world objects}
    \label{fig:testing_real}
  \end{subfigure}
  \label{fig:testing}
  \begin{subfigure}{0.485\textwidth}
  \includegraphics[width=\textwidth]{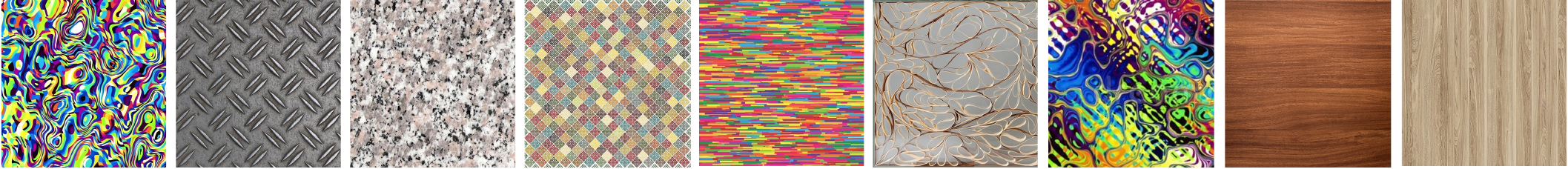}
  \vspace{-5mm}
    \caption{Training phase background textures}
    \label{fig:training_bg}
  \end{subfigure}
    \begin{subfigure}{0.485\textwidth}
  \includegraphics[width=\textwidth]{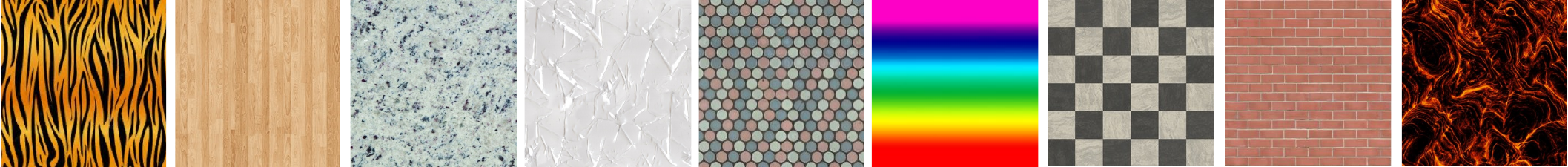}\
    \vspace{-5mm}
    \caption{Testing novel background textures from VIMA \cite{jiang2023vima}}
    \label{fig:testing_bg}
  \end{subfigure}
    \caption{Target objects used in both simulation and the real world include 32 instances for simulation and 15 for real-robot experiments. We apply domain randomization to the robot workspace during training for robust performances and use the textures from VIMA \cite{jiang2023vima} in testing.} 
    \vspace{-5mm}
\end{figure}

\noindent\textbf{Implementation Details:} Following the collection of the RGA dataset, both the OGRG-RGA grounding model and the MGN affordance prediction model are trained for 50 epochs, with batch sizes of 12 and 32, respectively. The AdamW optimizer is employed for the OGRG-RGA model, with an initial learning rate of $5 \times 10^{-5}$ and a polynomial learning rate decay. The maximum sentence length is set to 25 tokens. All RGA-related models are trained and tested on a single NVIDIA RTX 2080 Ti GPU.

\noindent\textbf{Evaluation Metrics:} The object grounding task is evaluated using the overall Intersection over Union (oIoU), similar to the metric used in RGS. For object grasping, the evaluation metric is defined as the object instance grasp success rate:
$\frac{\text{\# of successful grasps on the correct target}}{\text{\# of total grasps}}$.
In each test case, a single grasp attempt is executed. A grasp is considered successful only if the correct target object is grasped.

\noindent\textbf{OGRG-RGA Object Grounding in Simulation:} Test scenes were collected with varying numbers of objects, ranging from 1 to 7 per scene. For spatial attribute grounding experiments, the \textit{Abs} and \textit{Rel} settings correspond to absolute and relative spatial reasoning, respectively. A total of 1000 scenes were formulated, with and without object repetition (e.g., five objects consisting of three apples, one banana, and one tissue box). For general attribute grounding experiments, two language templates were used: \textit{Attr-cls}, which includes color, shape, and category names, and \textit{Attr-base}, which includes only color and shape attributes. These experiments were conducted on an additional 777 scenes.

The object grounding results are presented in Table \ref{tab:rga-ablation}. ETRG \cite{yu2024parameter} is used as a baseline, modified to predict grounding masks with minor adjustments to its architecture. LAVT \cite{yang2022lavt} uses a unidirectional aligner and no depth fusion; Variants of the OGRG-RGA model were also evaluated, OGRG-nodepth, which incorporates a bidirectional aligner without depth fusion; and OGRG-db, where the multimodal fused features $f_{back\_v}^i$ and $f_{back\_l}^i$ are passed directly to the next following aligner, similar to the depth branch in ETRG. The proposed OGRG-RGA model outperforms all baselines, achieving an average improvement of $+3.48\%$ in oIoU compared to ETRG.

\begin{table}[t]
    \centering
    \caption{\textbf{OGRG-RGA object grounding performance (oIoU) in simulation.}}
    \begin{tabular}{l|l|c|c|c|c}
        \hline
        Baselines & Abs & Rel & Attr-cls & Attr-base & Avg\\
        \hline
        ETRG \cite{yu2024parameter} & 93.67 & 84.35 & 92.85 & 91.28 & 90.54\\
        LAVT \cite{yang2022lavt} & 95.62 & 85.83 & 95.34 & 94.24 & 92.76\\
        OGRG-nodepth & 95.88 & 84.59 & 95.55 &	94.91 &	92.73\\
        OGRG-db & 96.49 & 85.88 & 96.51 & \textbf{95.69} & 93.64\\
        \textbf{OGRG (Ours)} & \textbf{97.00} & \textbf{87.05} & \textbf{96.55} & 95.49 & \textbf{94.02}\\
        \hline
    \end{tabular}
    \vspace{-2mm}
    \label{tab:rga-ablation}
\end{table}

\noindent\textbf{OGRG-RGA Robot Grasping in Simulation:} For testing, 1,600 test cases were created across 32 objects with pre-selected query language descriptions based on object attributes. In each test case, three identical object instances were randomly dropped into the workspace, and spatial language expressions were used to specify the target. For fair comparisons, all baselines were tested under identical scenes and language expressions.

\begin{figure}[t]
    \centering
    \begin{subfigure}{0.48\textwidth}
        {\includegraphics[width=0.24\textwidth]{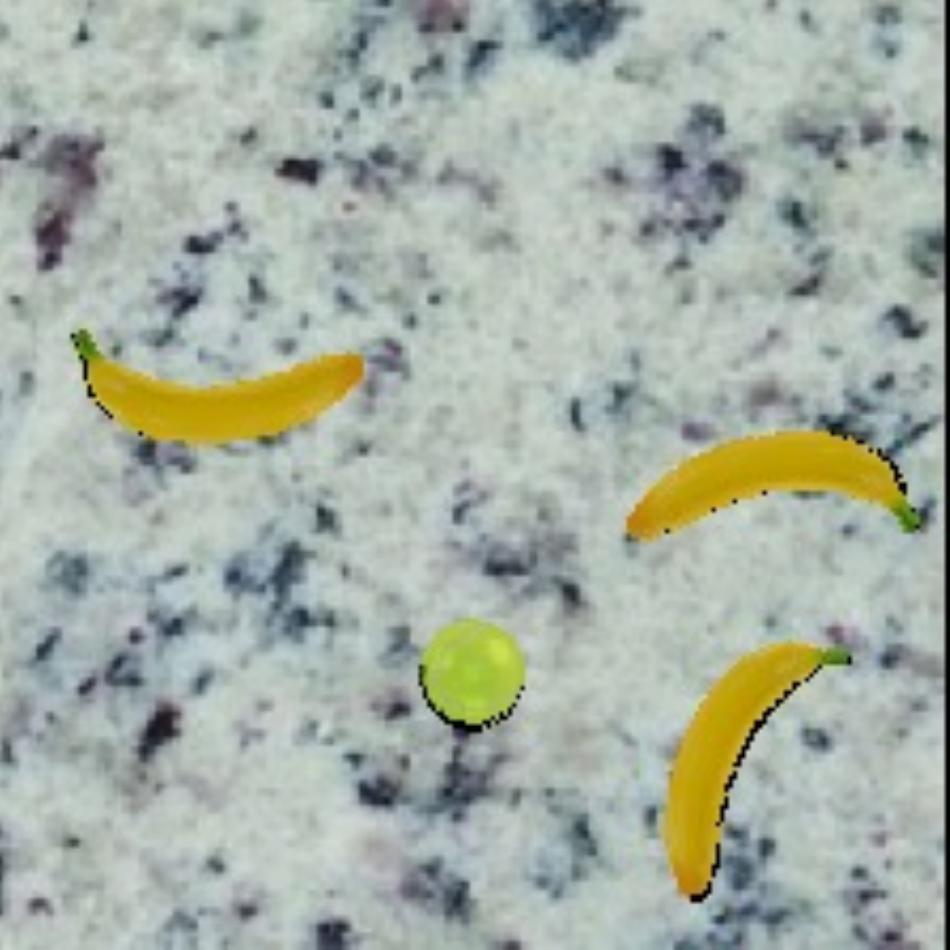}}\hfill
        {\includegraphics[width=0.24\textwidth]{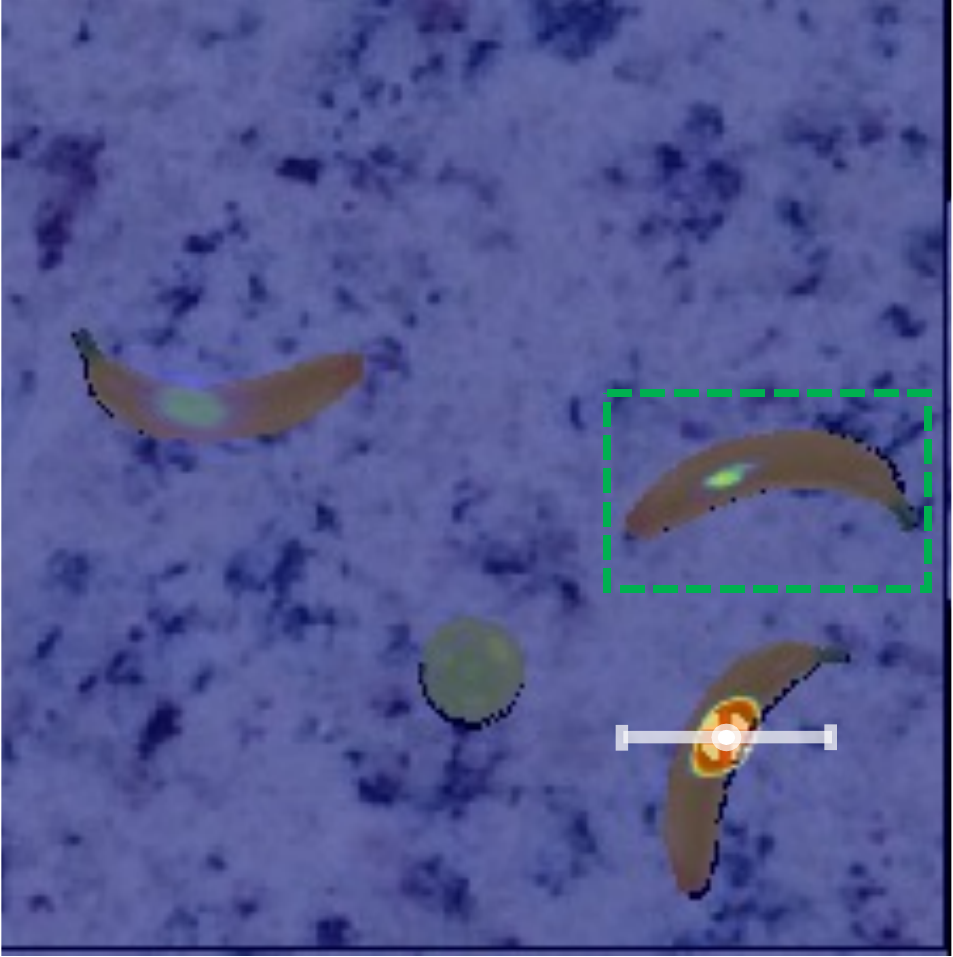}}\hfill
        {\includegraphics[width=0.24\textwidth]{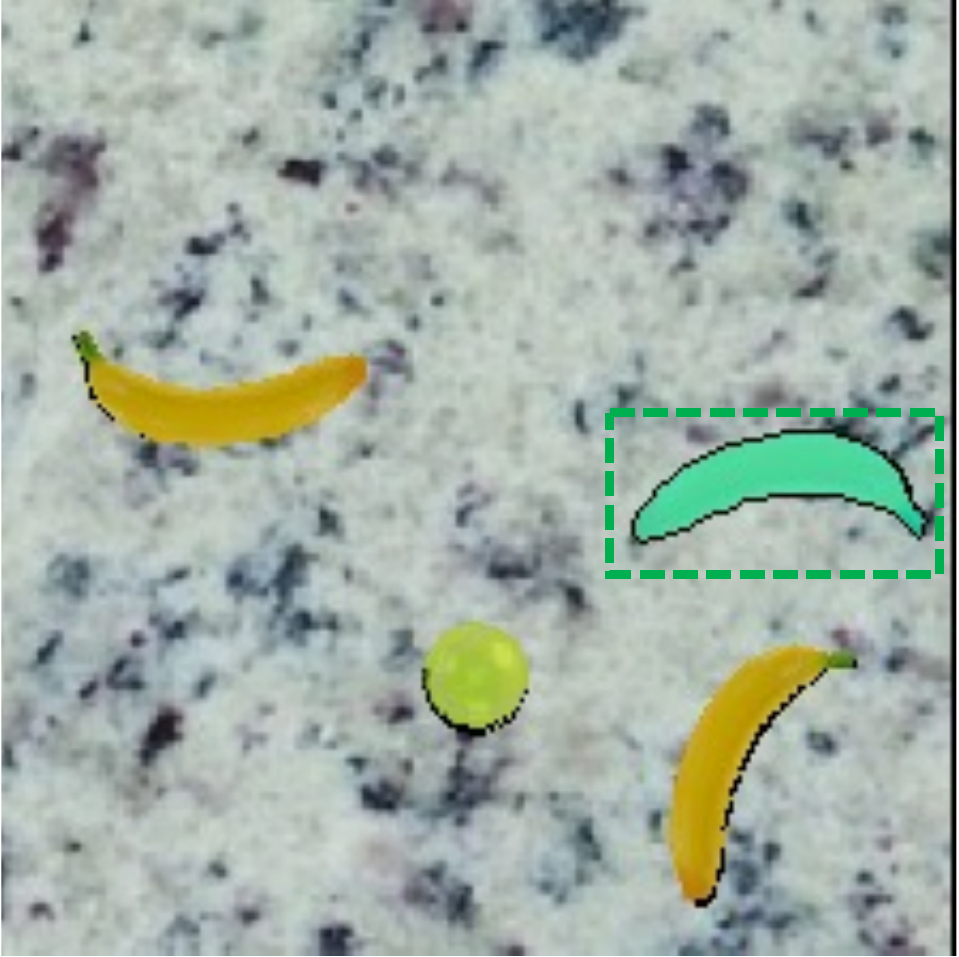}}\hfill
        {\includegraphics[width=0.24\textwidth]{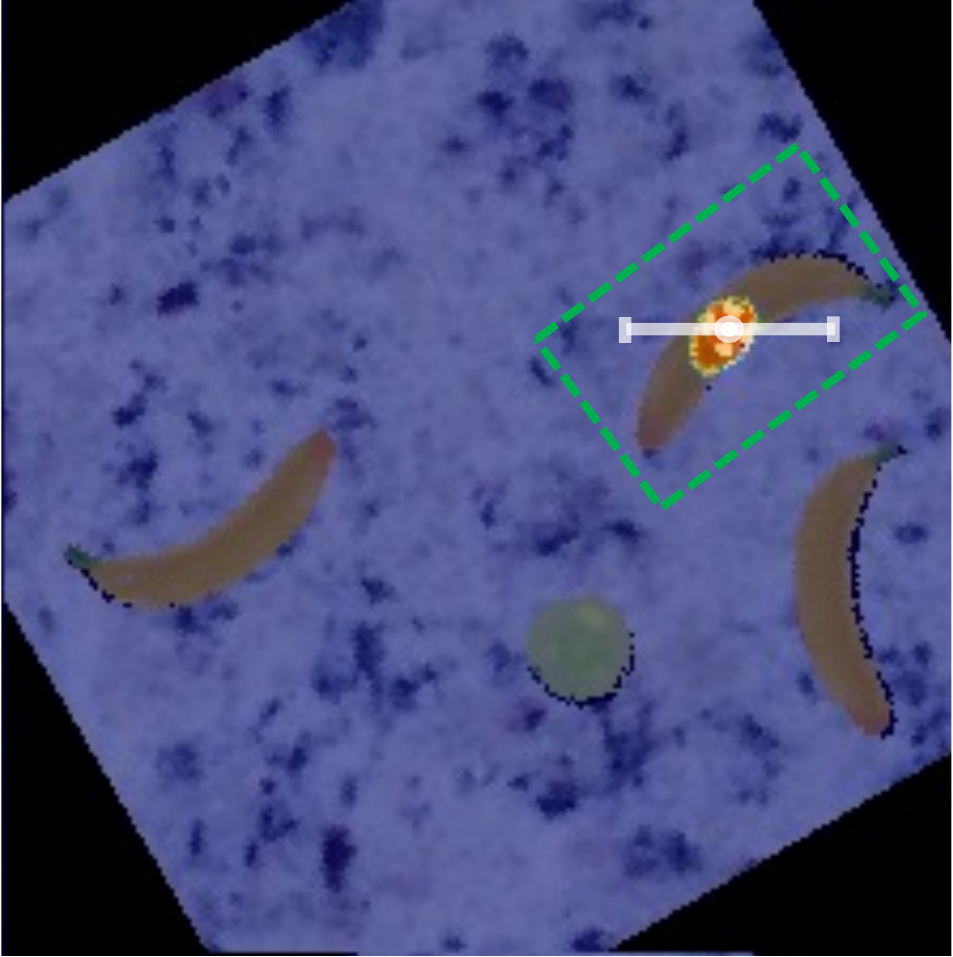}}%
        \vspace{-3pt}
        \caption{Language input: pass me the banana that is to the middle right of the workspace}
        \label{fig:4abs}
        \vspace{5pt}
    \end{subfigure}
    \begin{subfigure}{0.48\textwidth}
        {\includegraphics[width=0.24\textwidth]{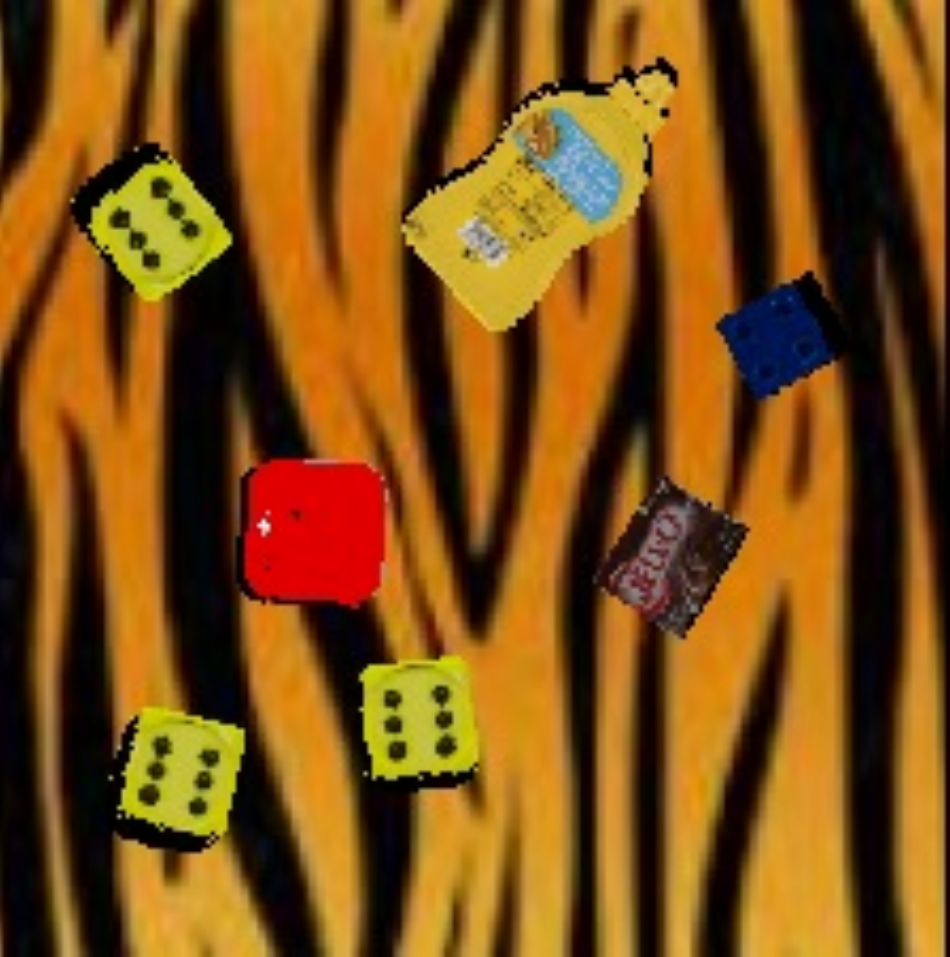}}\hfill
        {\includegraphics[width=0.24\textwidth]{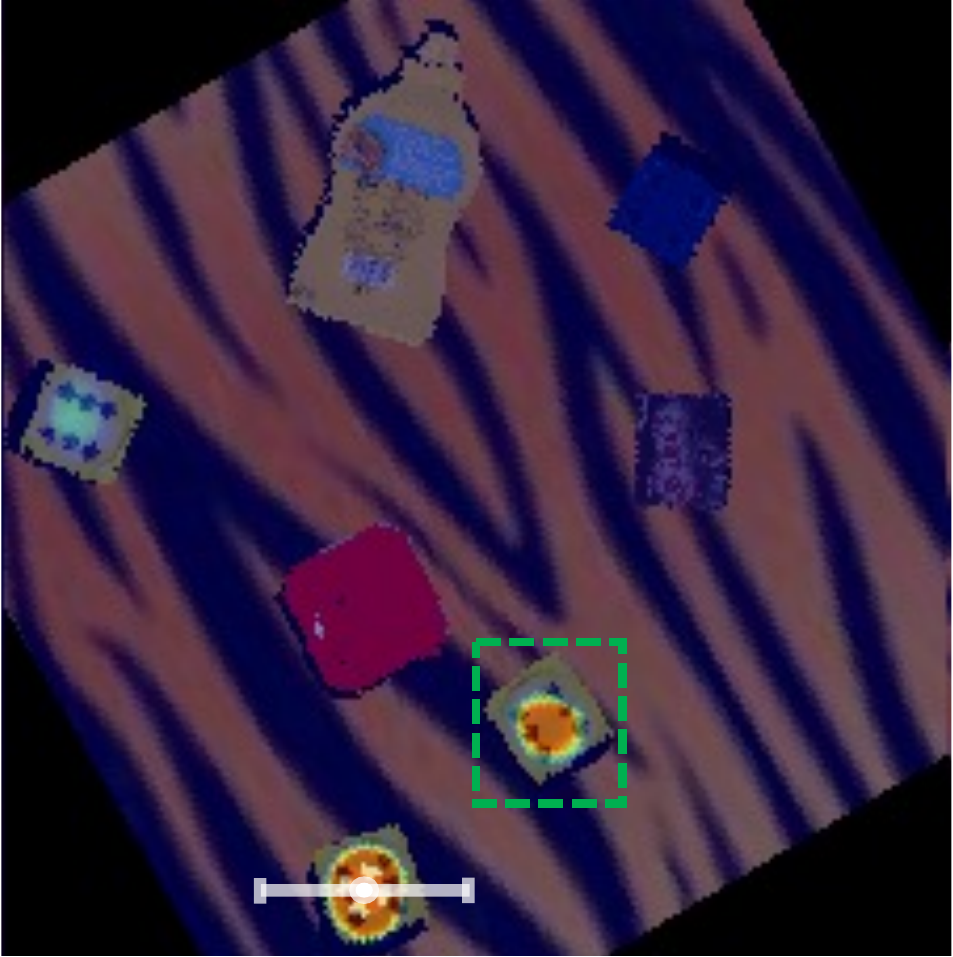}}\hfill
        {\includegraphics[width=0.24\textwidth]{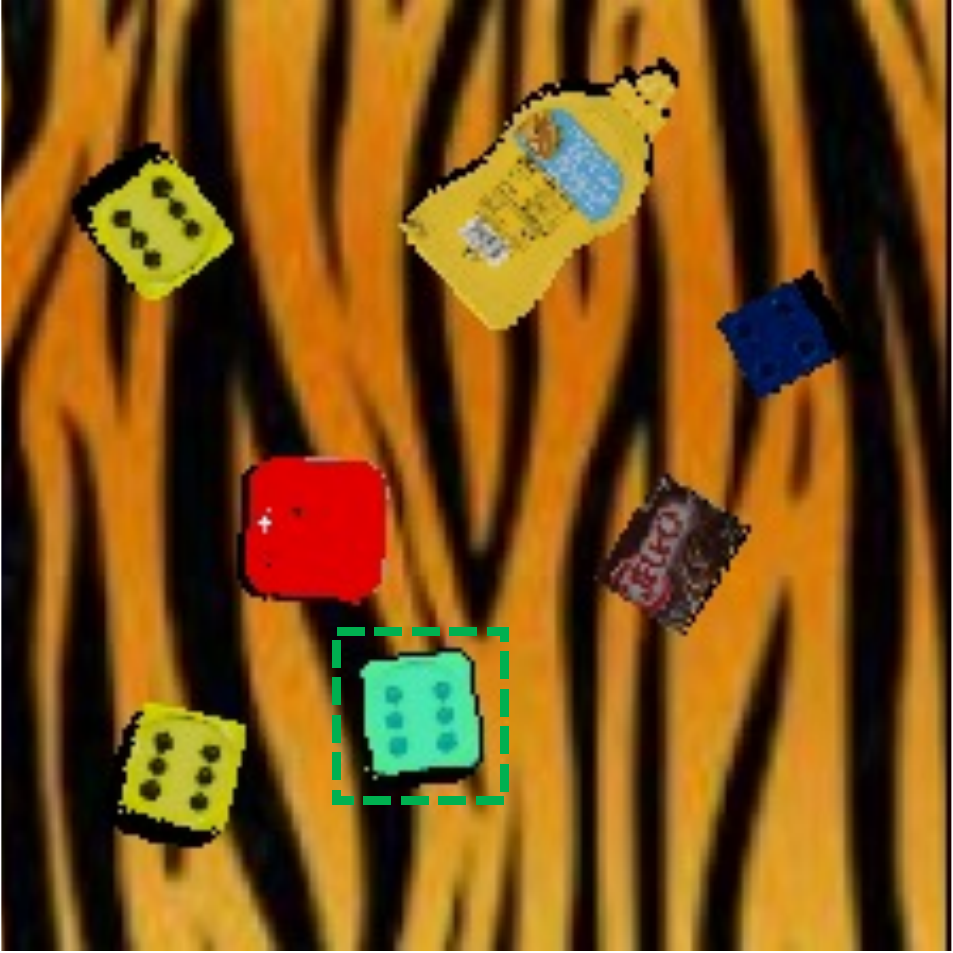}}\hfill
        {\includegraphics[width=0.24\textwidth]{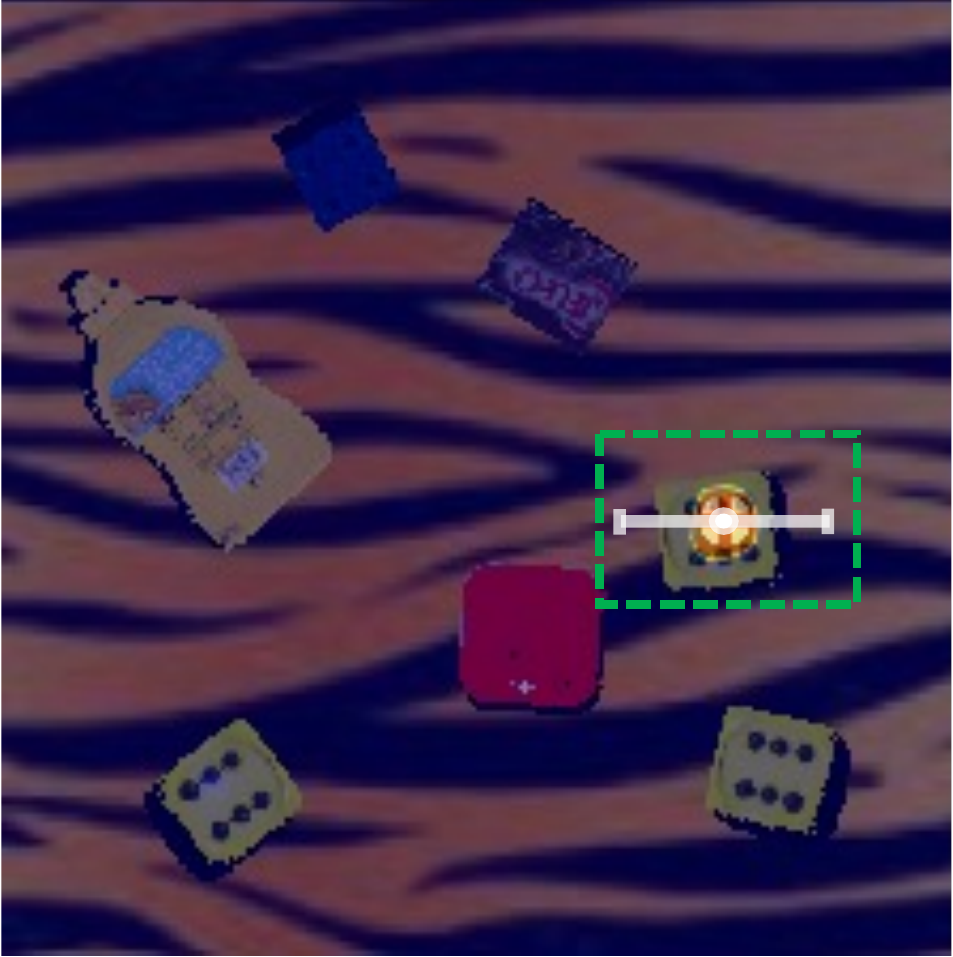}}%
        \vspace{-3pt}
        \caption{Language input: grasp me the bottom center dice}
        \label{fig:7abs}
        \vspace{5pt}
    \end{subfigure}
    \begin{subfigure}{0.48\textwidth}
        {\includegraphics[width=0.24\textwidth]{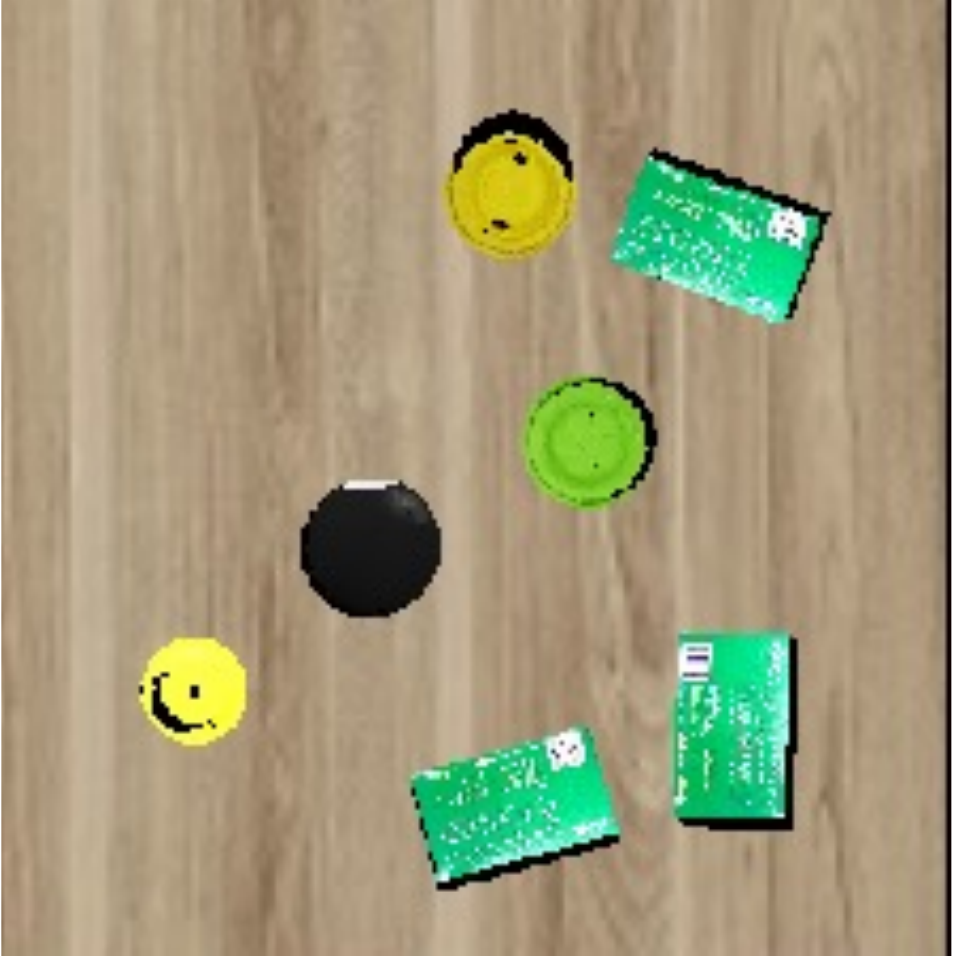}}\hfill
        {\includegraphics[width=0.24\textwidth]{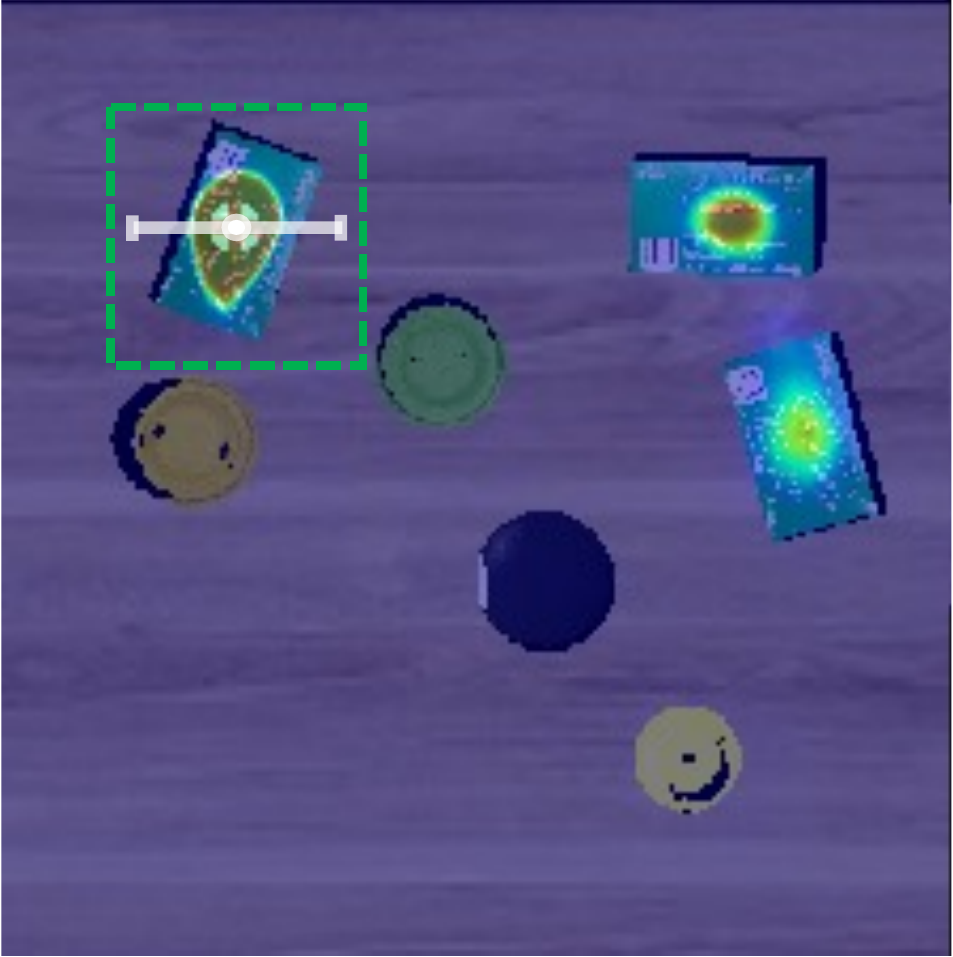}}\hfill
        {\includegraphics[width=0.24\textwidth]{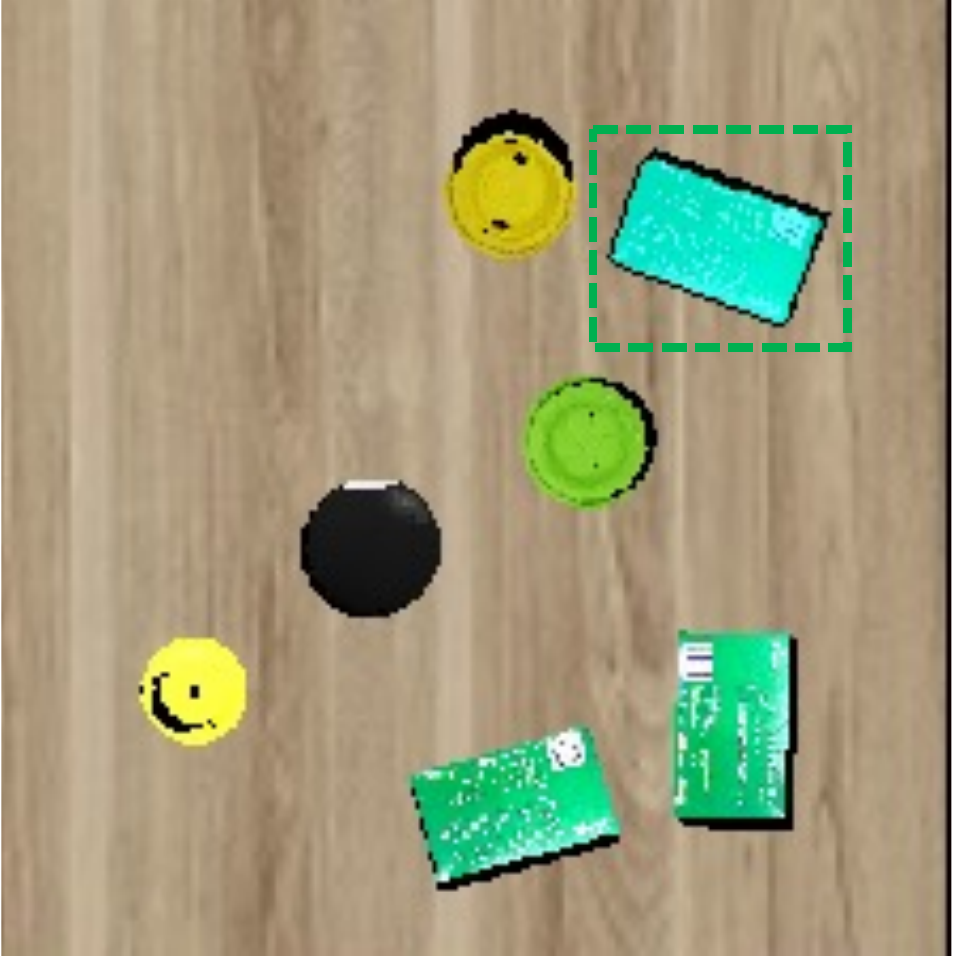}}\hfill
        {\includegraphics[width=0.24\textwidth]{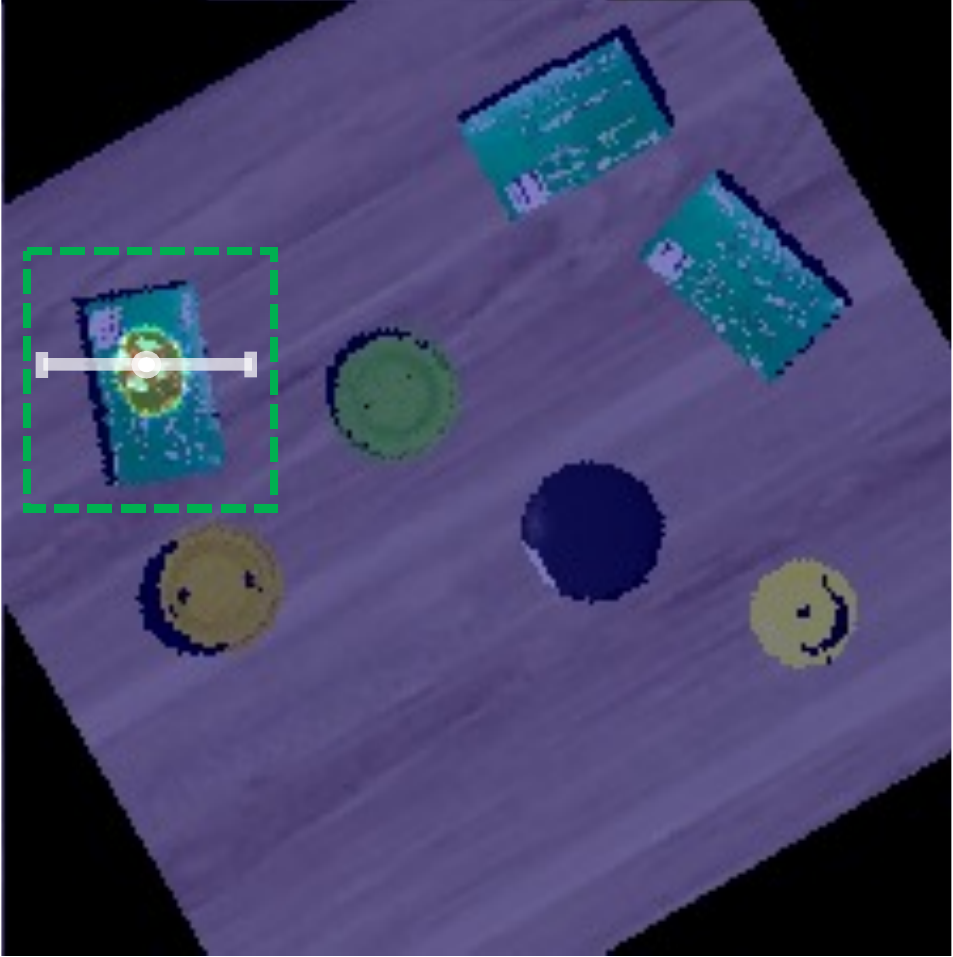}}%
        \vspace{-3pt}
        \caption{Language input: grasp the tissue box that is to the upper right of the green cylinder green cup}
        \label{fig:7rel}
        \vspace{-3pt}
    \end{subfigure}
    \caption{\textbf{Grounding masks and grasp affordances with spatial reasoning in simulation.} The green bounding boxes highlight the correct language-referred target object. The first column shows the input scene. The second column shows the affordance predictions from ETRG~\cite{yu2024parameter}. The third and fourth columns denote the grounding mask and grasp affordances from OGRG pipeline.}
    \label{fig:sim}
    \vspace{-5pt}
\end{figure}

The robot grasping performance results are presented in Table \ref{tab:rga-grasp}. The OGRG-RGA method significantly outperforms the state-of-the-art ETRG \cite{yu2024parameter} grasp affordance method, achieving a $+4.86\%$ improvement in overall grasping performance on the spatial reasoning task. Detailed qualitative visualizations are shown in Fig. \ref{fig:sim}. As demonstrated, the ETRG method in Fig. \ref{fig:4abs} and Fig. \ref{fig:7abs} fails to localize the target and provide successful grasp poses, while the proposed OGRG-RGA accurately segments the language-referred target and predicts feasible grasp poses. Furthermore, the grasp affordance maps generated by ETRG (Fig. \ref{fig:7rel}) exhibit redundant high values on incorrect object candidates, whereas the proposed approach focuses directly on the correct target object.


\begin{table}[t]
    \centering
    \caption{\textbf{Grasp-success rates of OGRG-RGA in simulation}, reported for seen and unseen background (BG) conditions to demonstrate generalization capability.}
    \begin{tabular}{l|l|c|c|c|c}
        \hline
        \multirow{2}{*}{BG} & \multirow{2}{*}{Baselines} & \multicolumn{2}{c|}{Absolute} & \multicolumn{1}{c|}{Relative} & \multirow{2}{*}{AVG} \\
        \cline{3-5}
         & & 4-obj & 7-obj & 7-obj & \\
        \hline
        \multirow{3}{*}{Seen} & ETRG~\cite{yu2024parameter} & 90.75 & 88.38 & 86.56 & 88.56 \\
         & OGRG-nodepth & 96.31 & 95.75 & 80.38 & 90.81 \\
         & \textbf{OGRG (Ours)} & \textbf{96.50} & \textbf{96.88} & \textbf{86.88} & \textbf{93.42} \\
        \hline
        \multirow{2}{*}{Unseen} & ETRG & 92.06 & 91.00 & \textbf{88.56} & 90.54 \\
         & \textbf{OGRG (Ours)} & \textbf{97.06} & \textbf{95.86} & 82.75 & \textbf{91.90} \\
        \hline
    \end{tabular}
    \label{tab:rga-grasp}
\end{table}

\begin{figure}[t]
    \centering
    \begin{subfigure}{0.48\textwidth}
        {\includegraphics[width=0.24\textwidth]{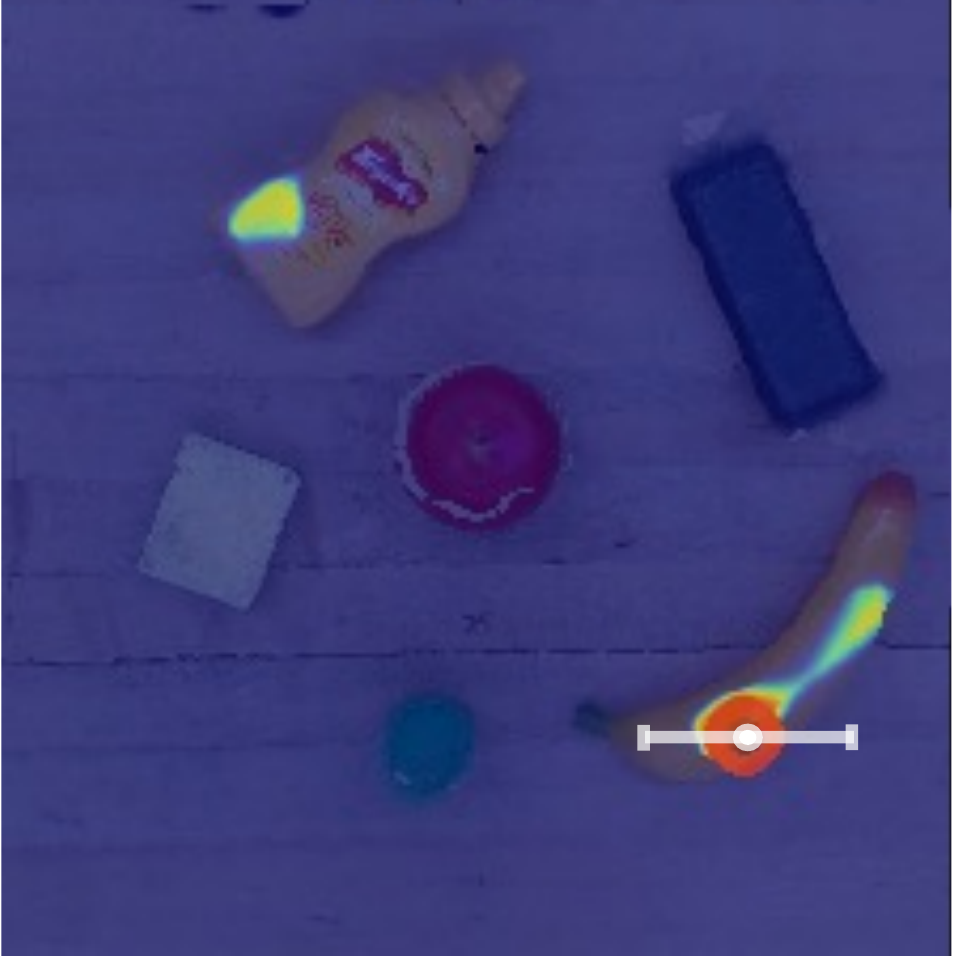}}\hfill
        {\includegraphics[width=0.24\textwidth]{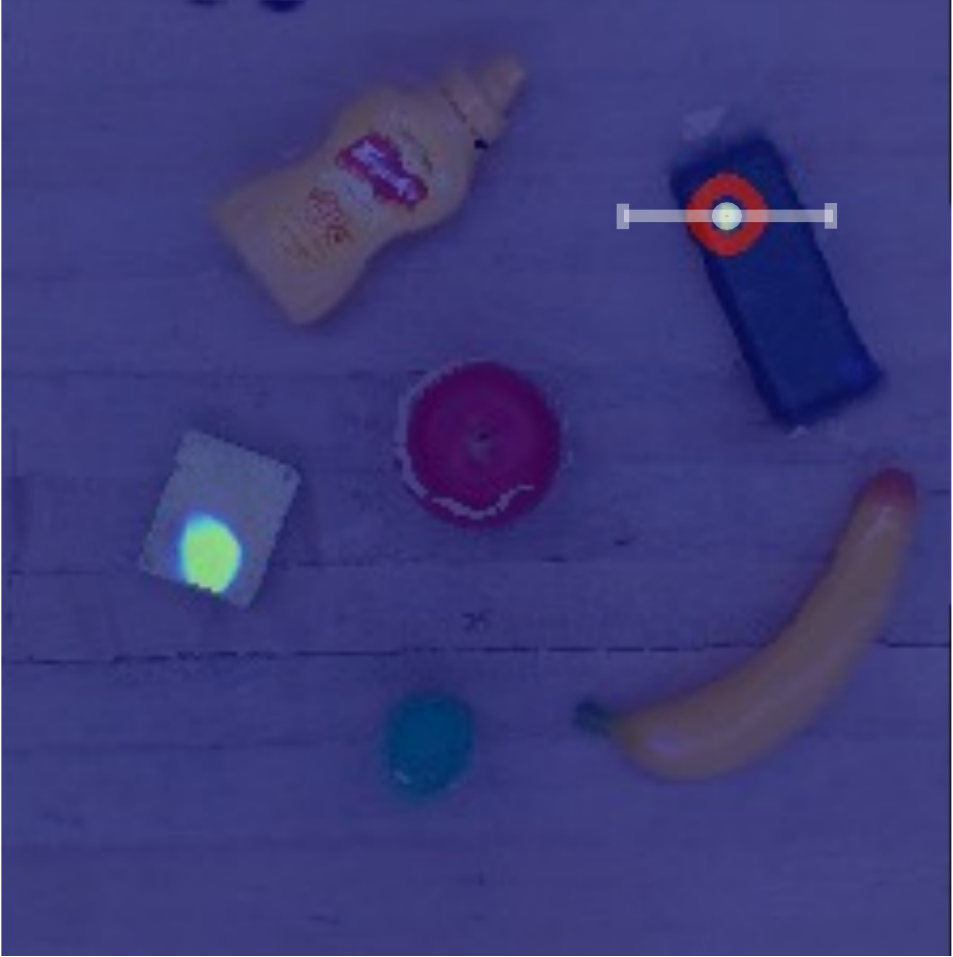}}\hfill
        {\includegraphics[width=0.24\textwidth]{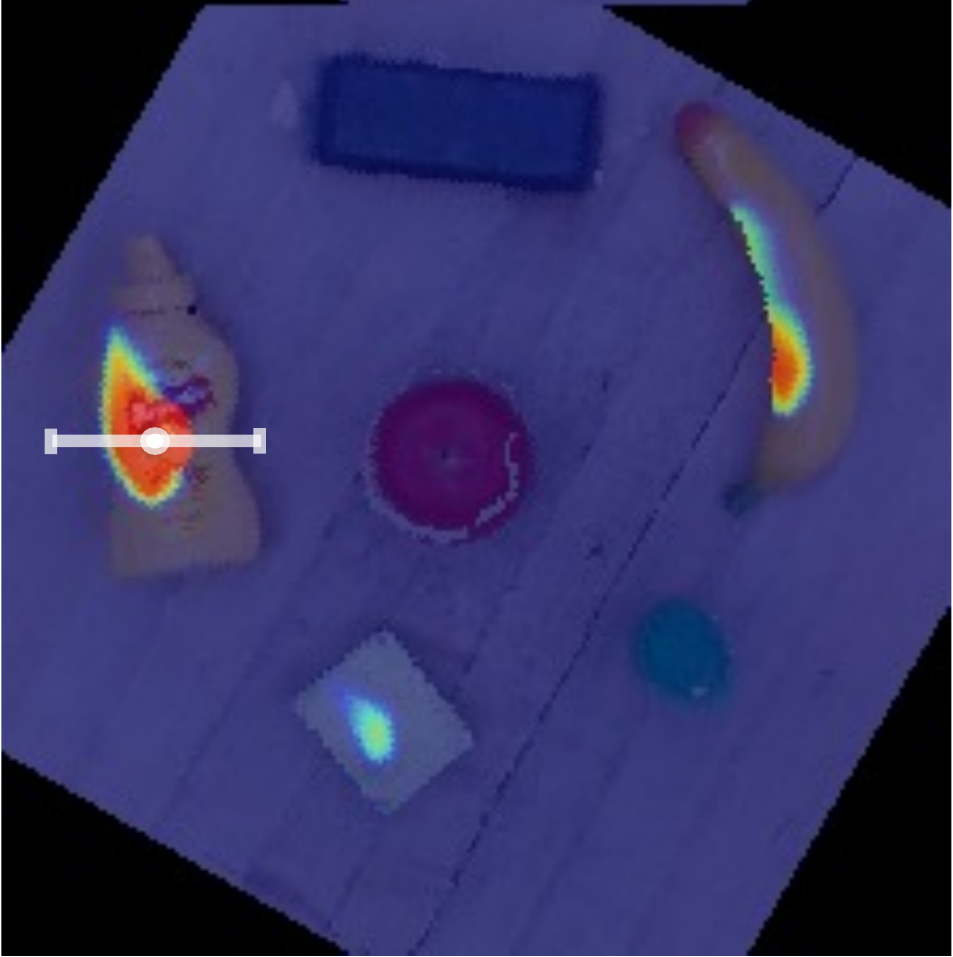}}\hfill
        {\includegraphics[width=0.24\textwidth]{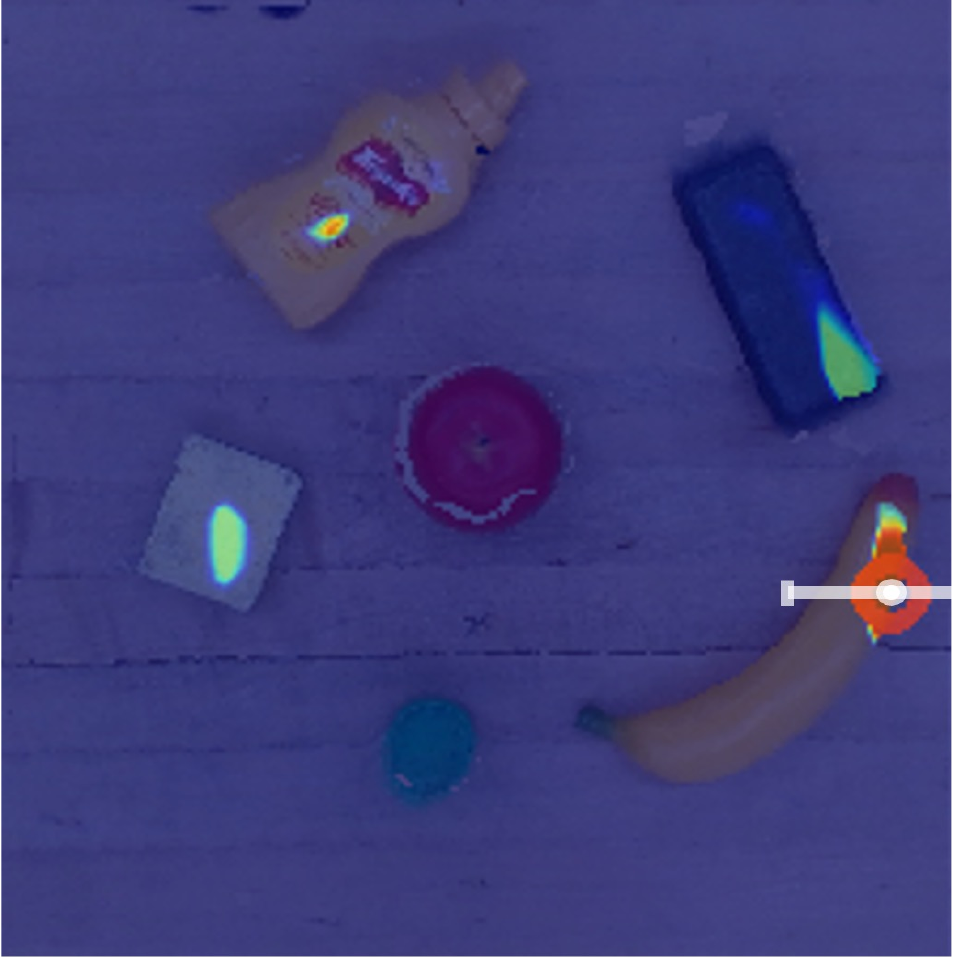}}%
        \vspace{-3pt}
        \caption{Grasp affordances \textit{Attribute-Grasp} \cite{yang2024attribute}}
        \label{fig:attrg_ns}
        \vspace{5pt}
    \end{subfigure}
    \begin{subfigure}{0.48\textwidth}
        {\includegraphics[width=0.24\textwidth]{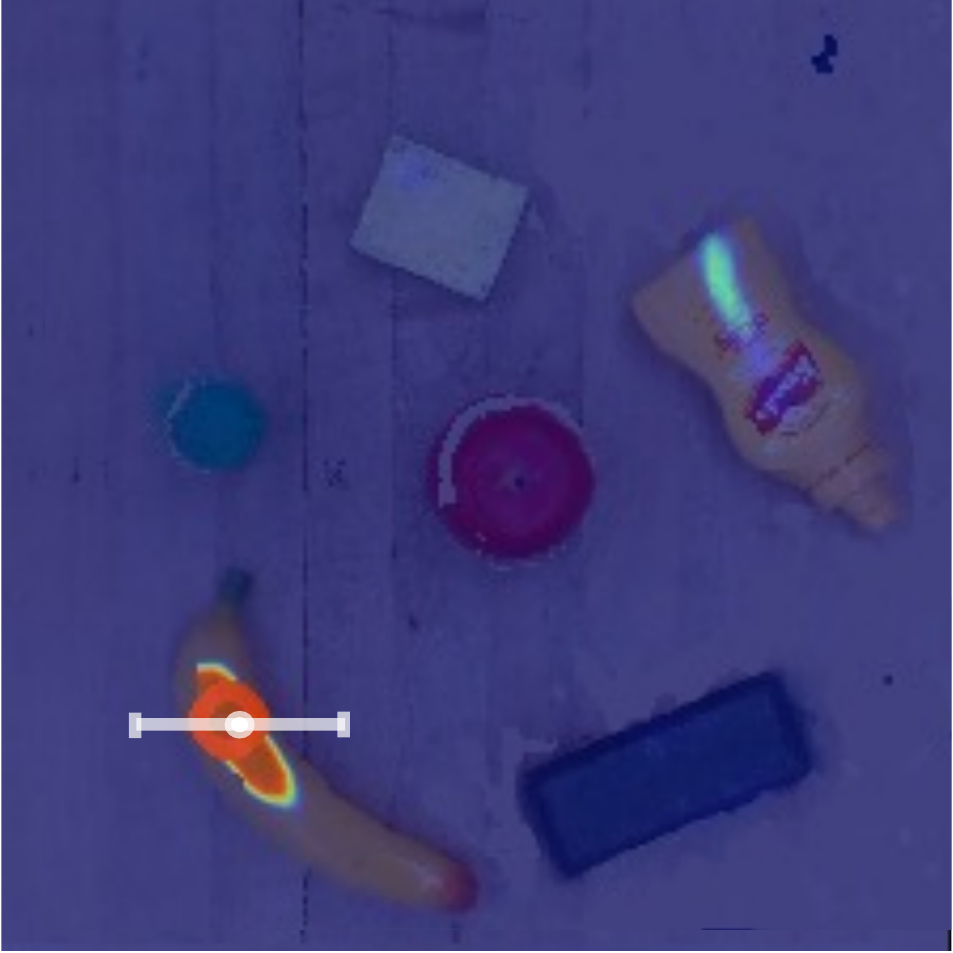}}\hfill
        {\includegraphics[width=0.24\textwidth]{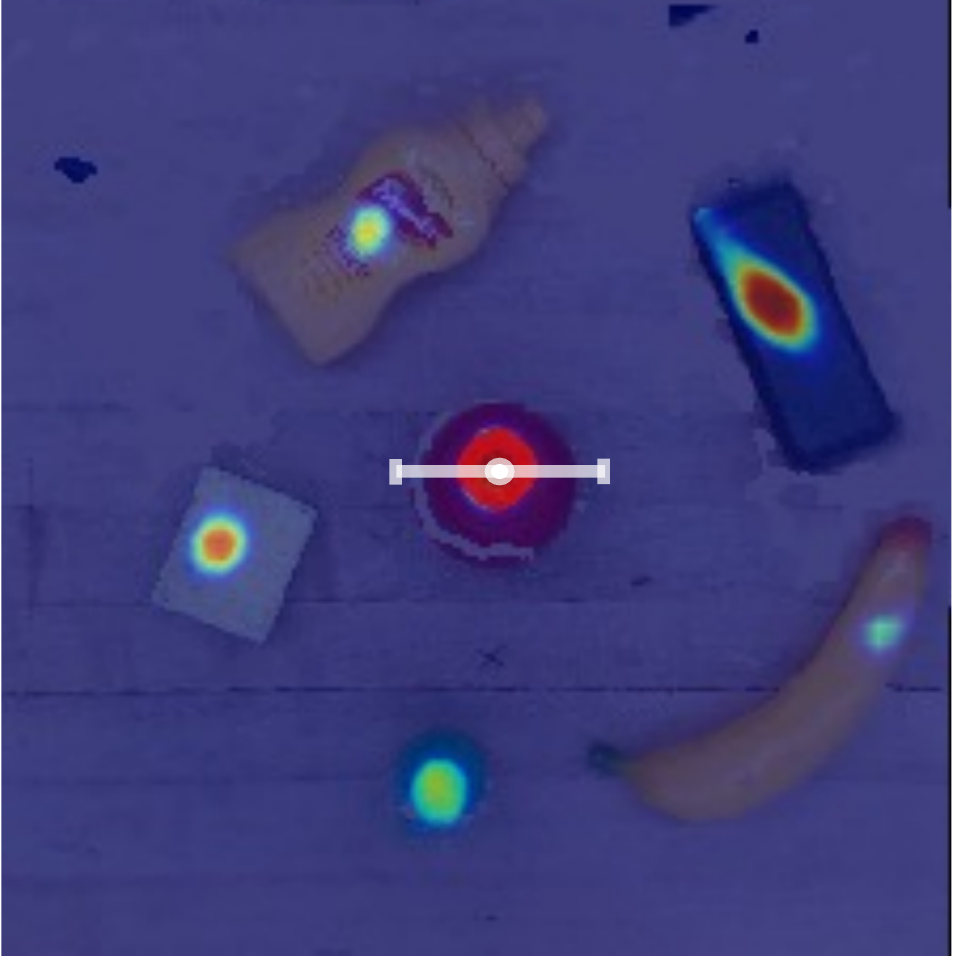}}\hfill
        {\includegraphics[width=0.24\textwidth]{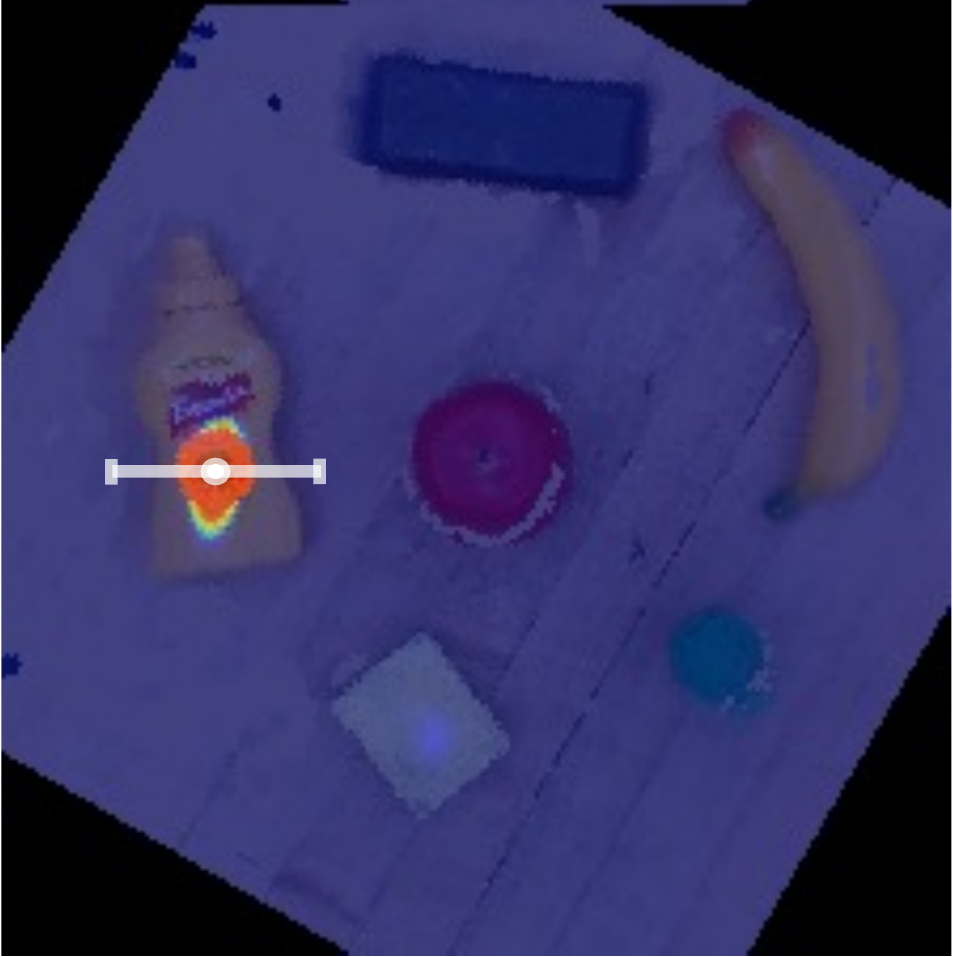}}\hfill
        {\includegraphics[width=0.24\textwidth]{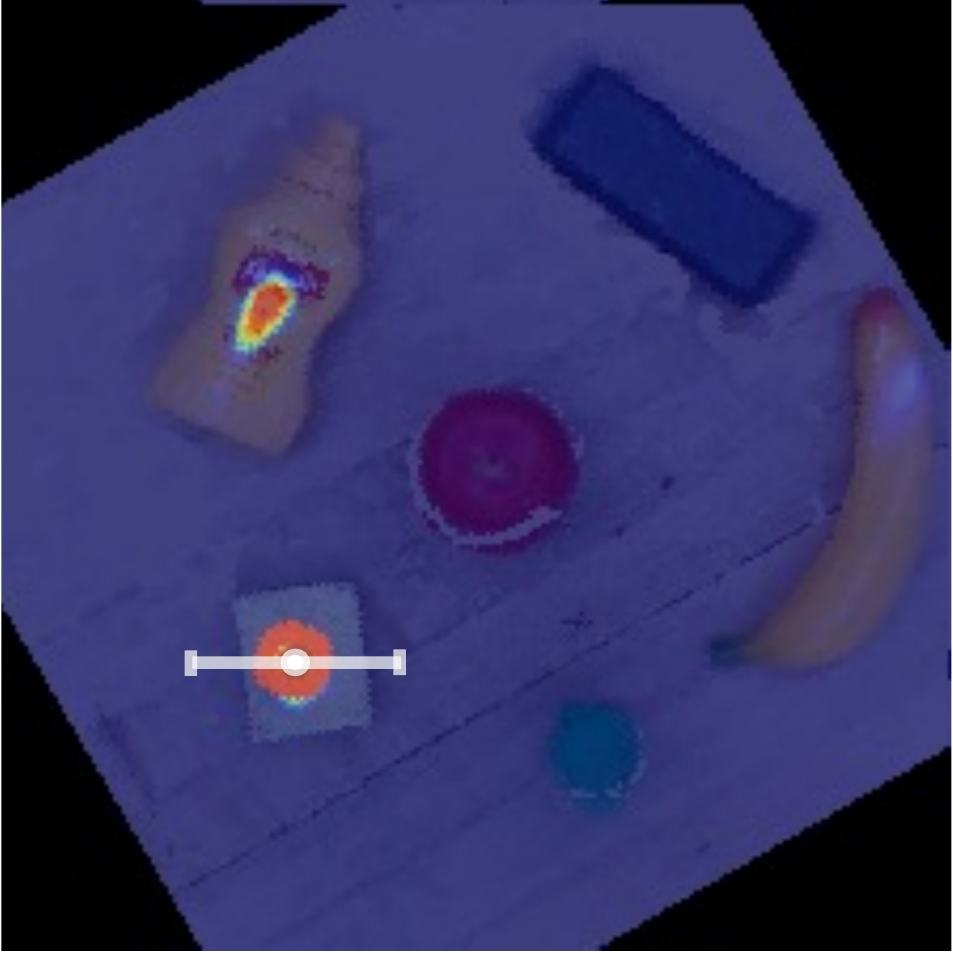}}%
        \vspace{-3pt}
        \caption{Grasp affordances \textit{ETRG} \cite{yu2024parameter}}
        \label{fig:ETRG_ns}
        \vspace{5pt}
    \end{subfigure}
    \begin{subfigure}{0.48\textwidth}
        {\includegraphics[width=0.24\textwidth]{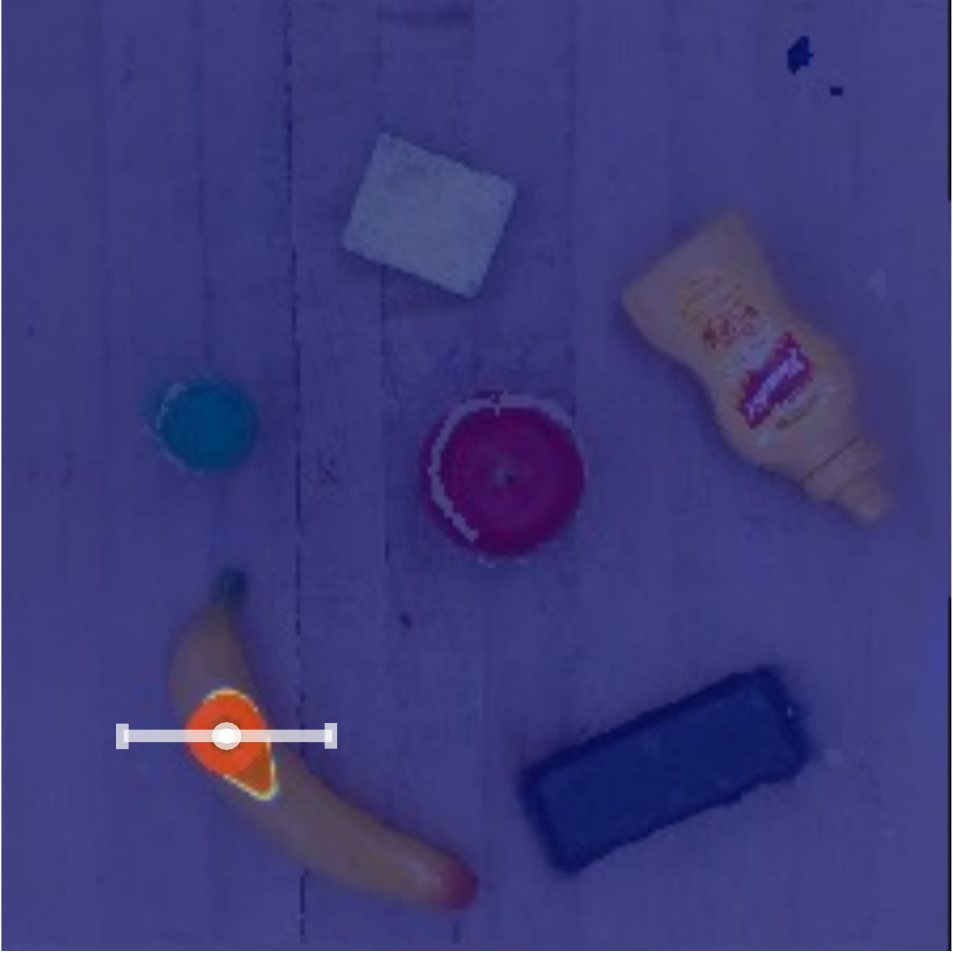}}\hfill
        {\includegraphics[width=0.24\textwidth]{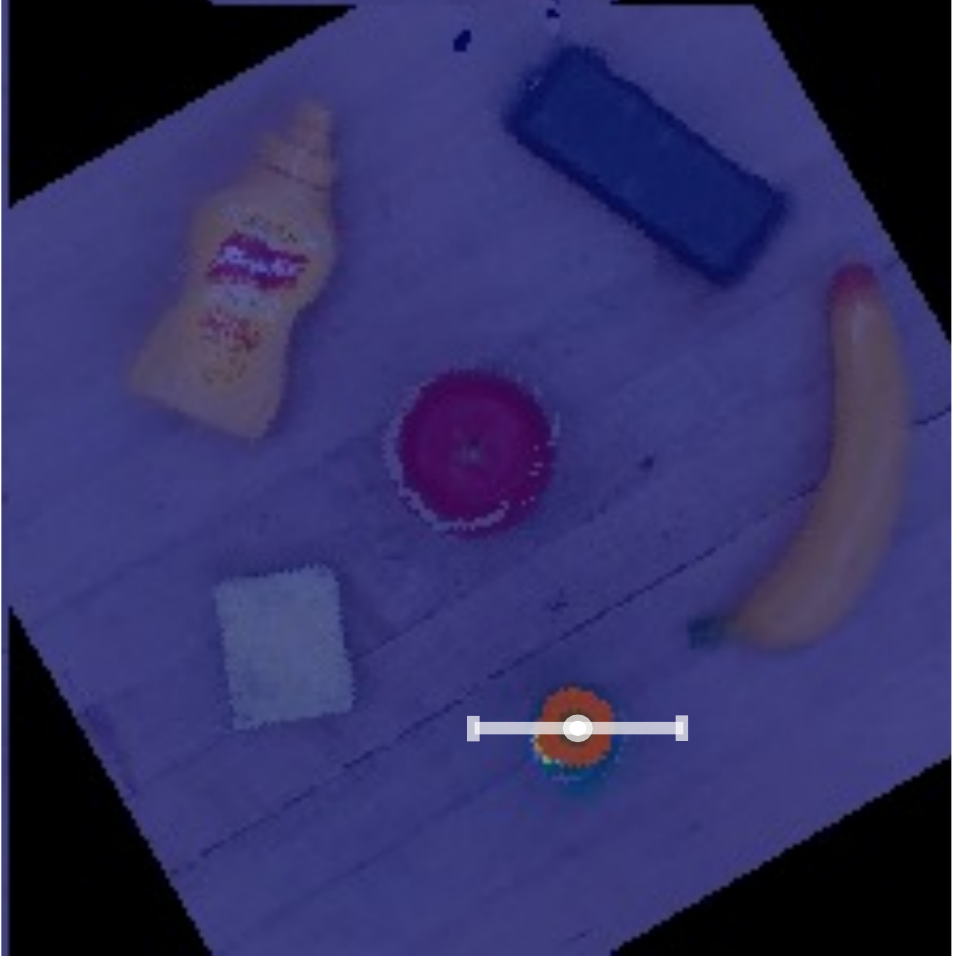}}\hfill
        {\includegraphics[width=0.24\textwidth]{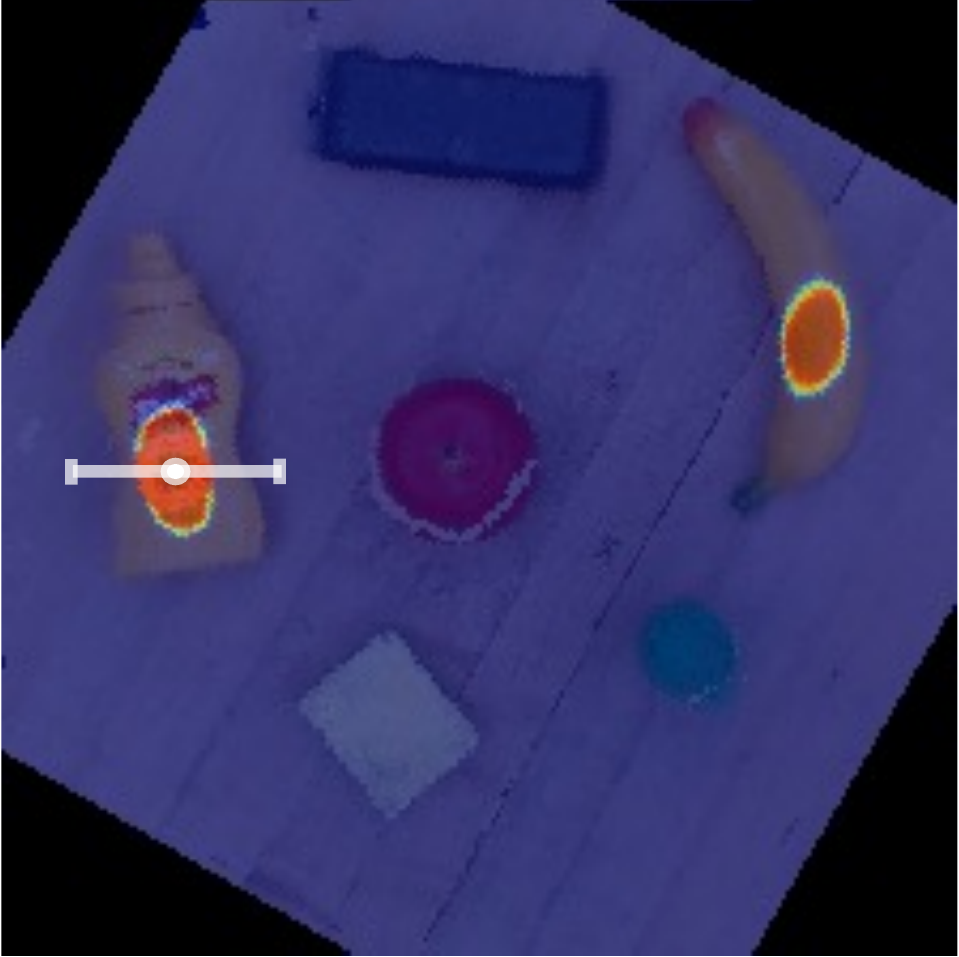}}\hfill
        {\includegraphics[width=0.24\textwidth]{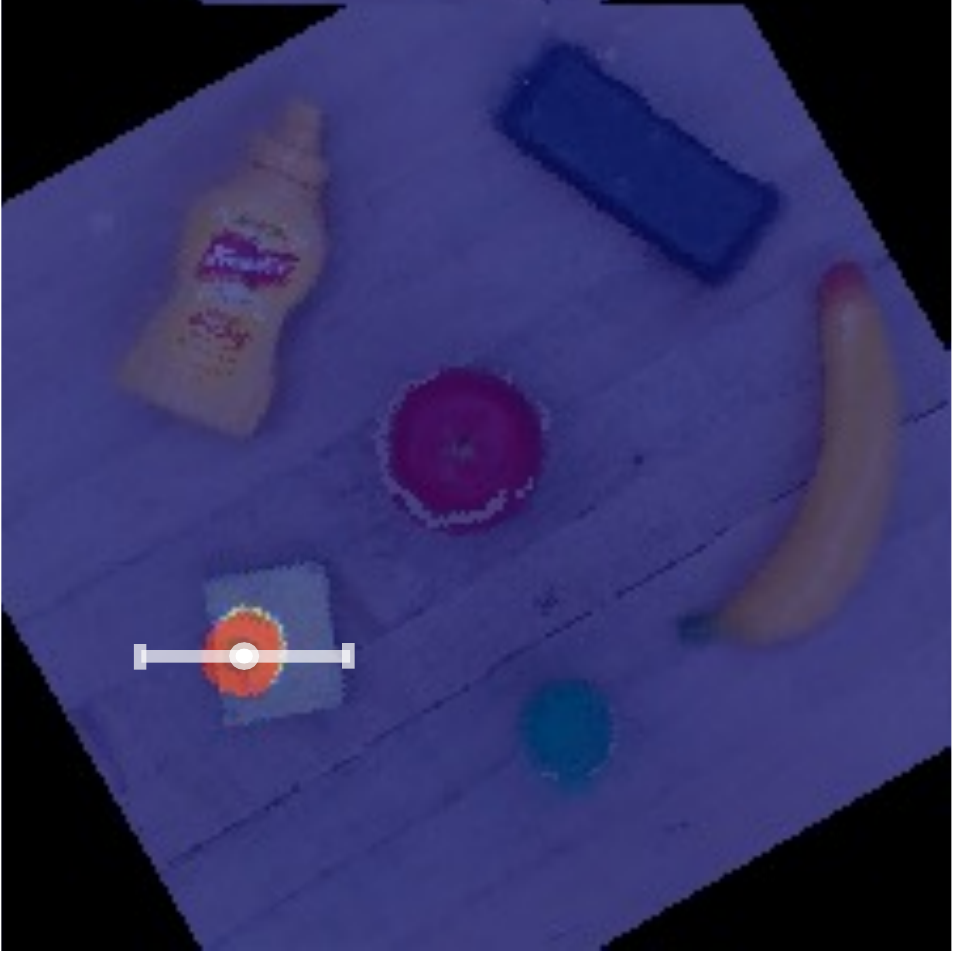}}%
        \vspace{-3pt}
        \caption{Grasp affordances \textit{OGRG-{nodepth}}}
        \label{fig:OGRG_nd_ns}
        \vspace{5pt}
    \end{subfigure}
    \begin{subfigure}{0.48\textwidth}
        {\includegraphics[width=0.24\textwidth]{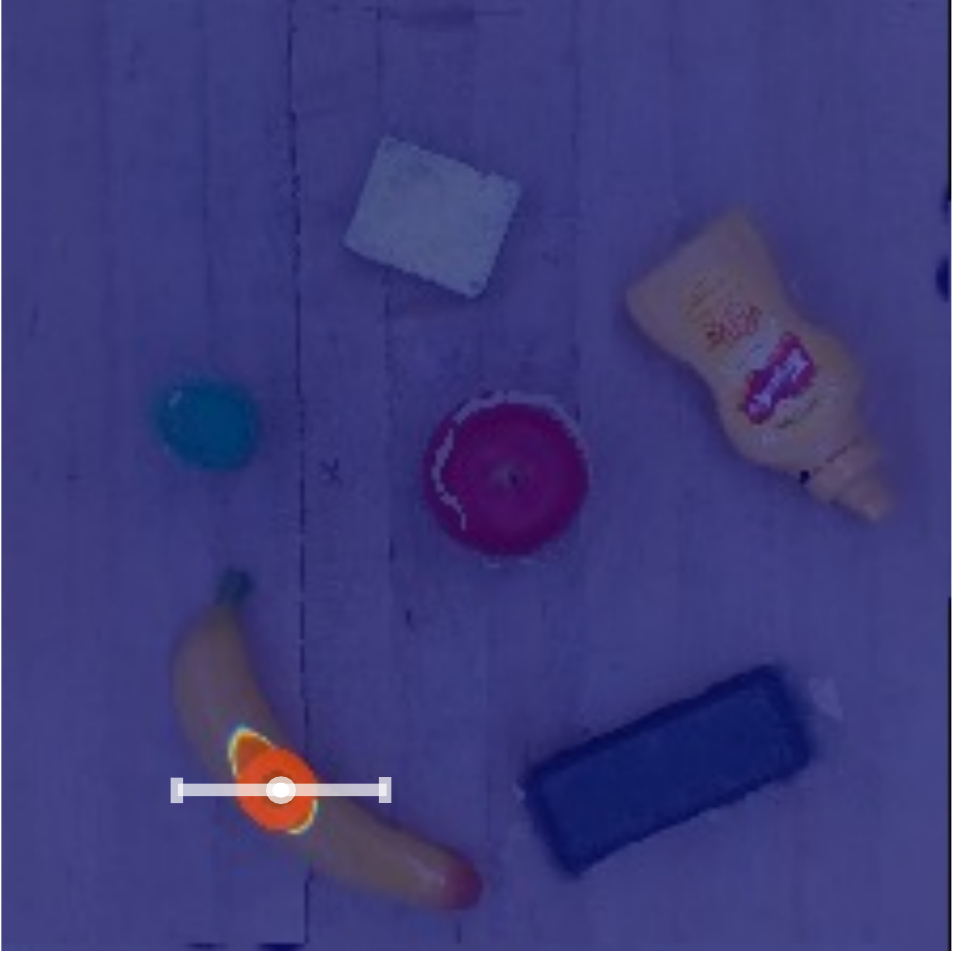}}\hfill
        {\includegraphics[width=0.24\textwidth]{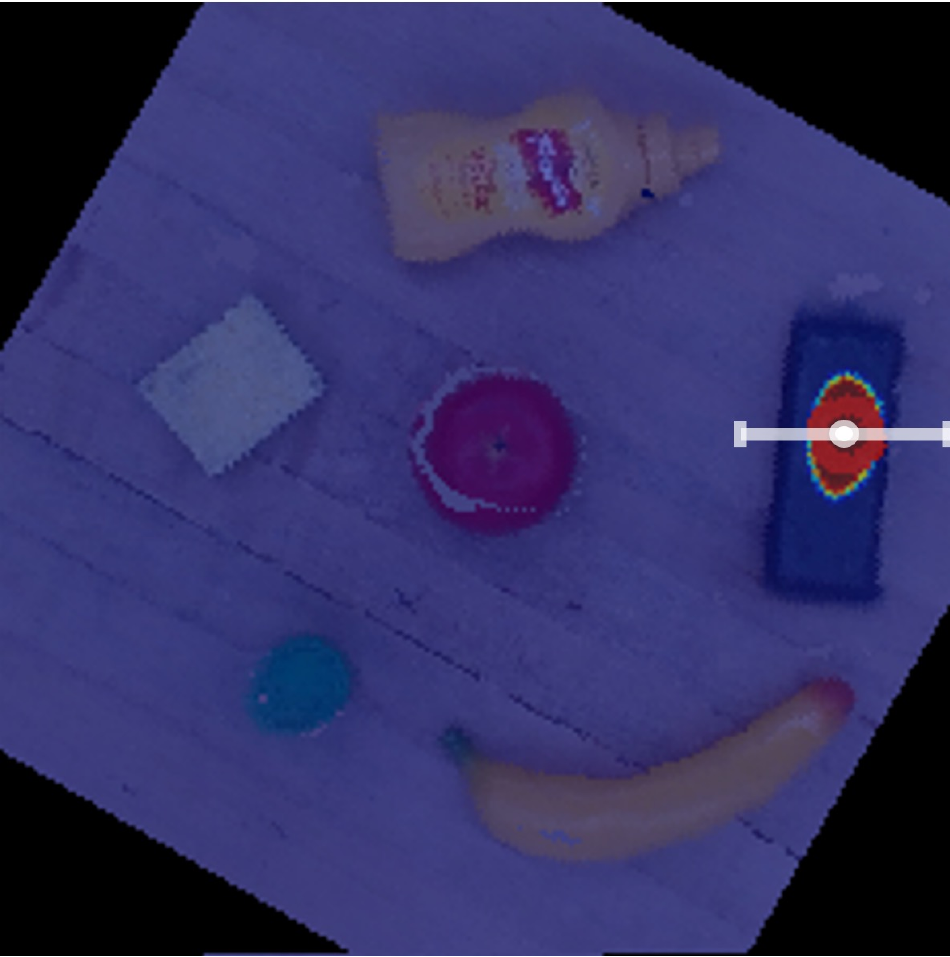}}\hfill
        {\includegraphics[width=0.24\textwidth]{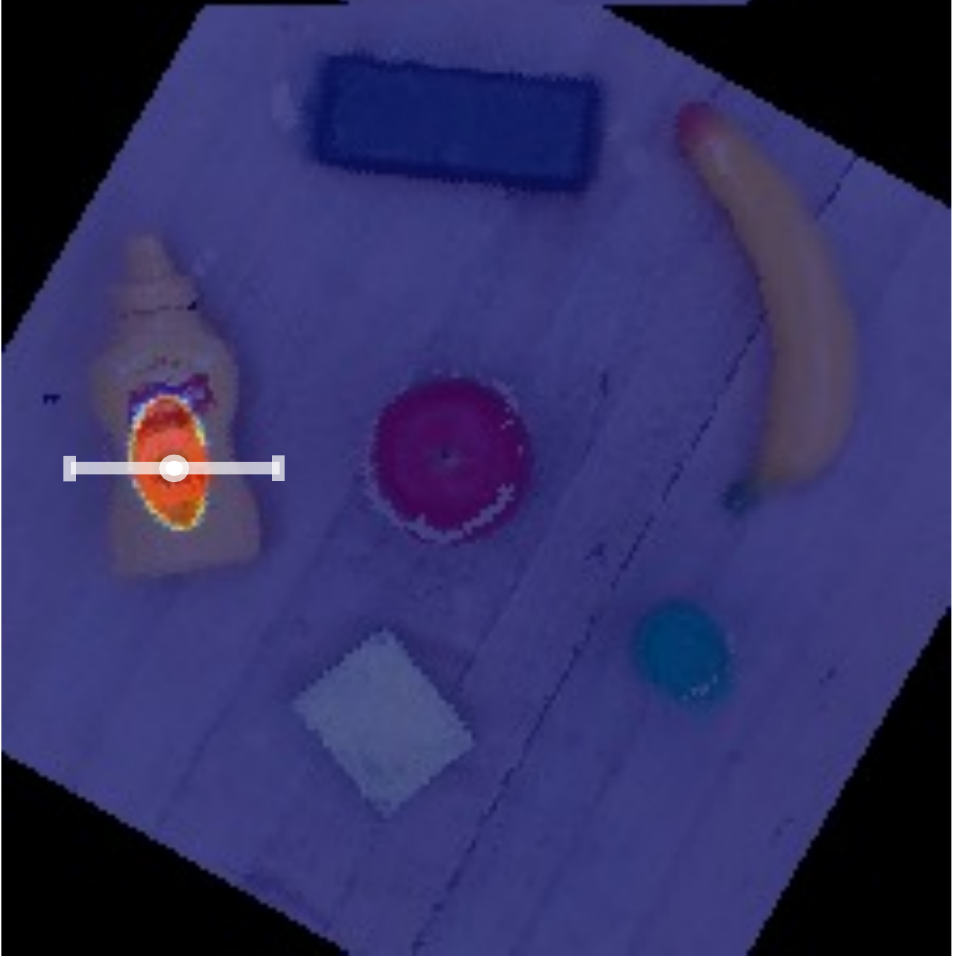}}\hfill
        {\includegraphics[width=0.24\textwidth]{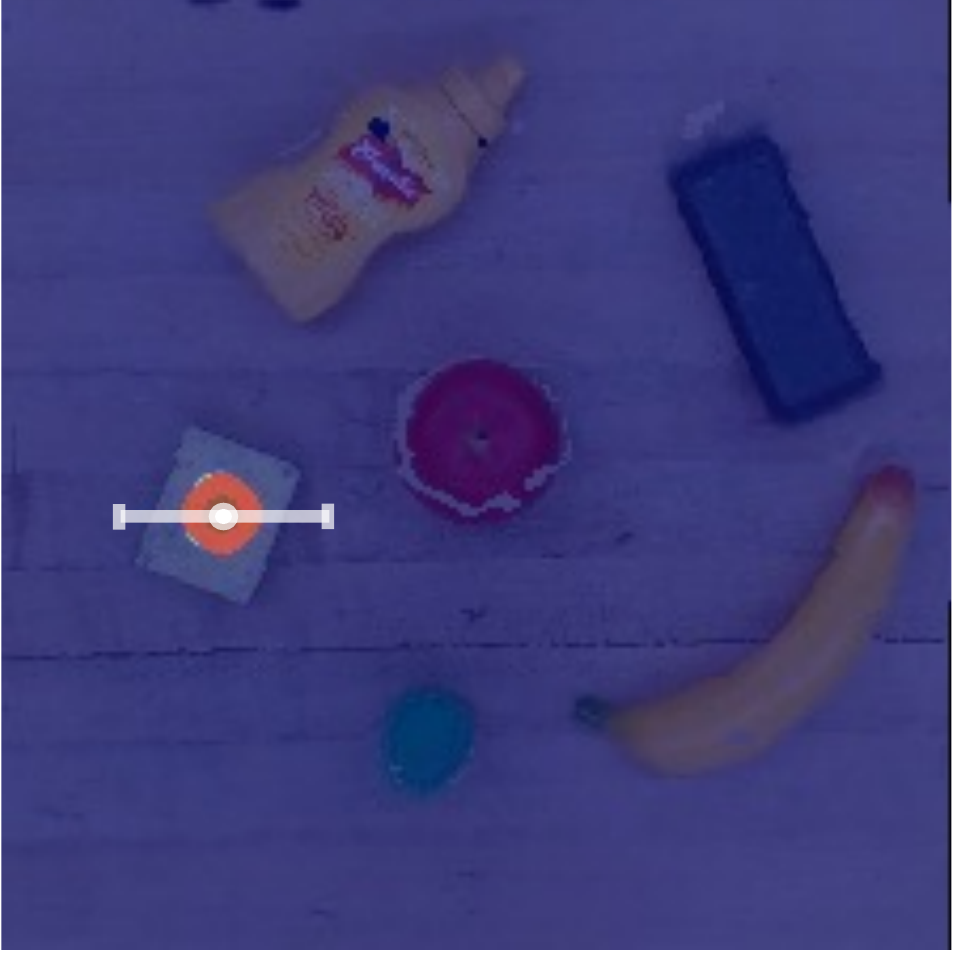}}%
        \vspace{-3pt}
        \caption{Grasp affordances \textit{OGRG \textbf{(Ours)}}}
        \label{fig:OGRG_ns}
    \end{subfigure}
    \caption{\textbf{Visualization of grasping affordances for real robot experiments with distinct challenging objects.} (\subref{fig:attrg_ns}) to (\subref{fig:OGRG_ns}) show the grasp affordances maps from the RGA models: \textit{Attribute-Grasp} \cite{yang2024attribute}, \textit{ETRG} \cite{yu2024parameter}, \textit{OGRG-{nodepth}}, and \textit{OGRG (Ours)}, respectively. The language inputs from left to right columns are: yellow banana, black cuboid, mustard bottle, and yellow cuboid sponge.}
    \label{fig:real-ns}
    \vspace{-5pt}
\end{figure}

\noindent\textbf{OGRG-RGA Real Robot Grasping:} Real robot experiments were conducted using a Franka Emika Panda robot equipped with a FESTO DHAS soft gripper. As shown in Fig. \ref{fig:testing_real}, 15 household objects were collected, and testing scenes were created by randomly sampling 6 target objects. A total of 100 visual-language-grasp triplets were manually collected in the real robot setup, and 246 grasping data points were generated after applying data augmentation \cite{yang2024attribute}. For each baseline, all models were fine-tuned using the same augmented dataset. Comparing with ETRG \cite{yu2024parameter} test scene setup, we parsed object candidates that have similar object attributes formulating more challenging real-robot scenes for open-form language comprehension.

\begin{figure}[ht]
    \centering
    \begin{subfigure}{0.48\textwidth}
        {\includegraphics[width=0.24\textwidth]{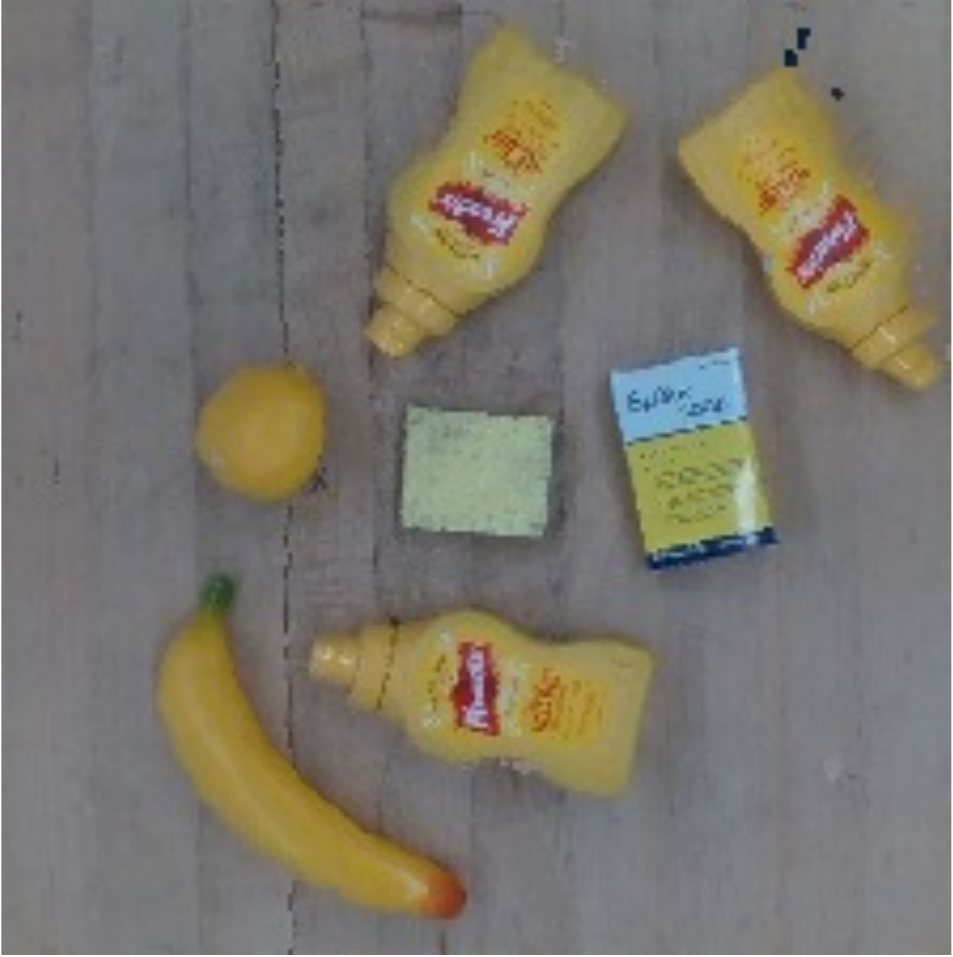}}\hfill
        {\includegraphics[width=0.24\textwidth]{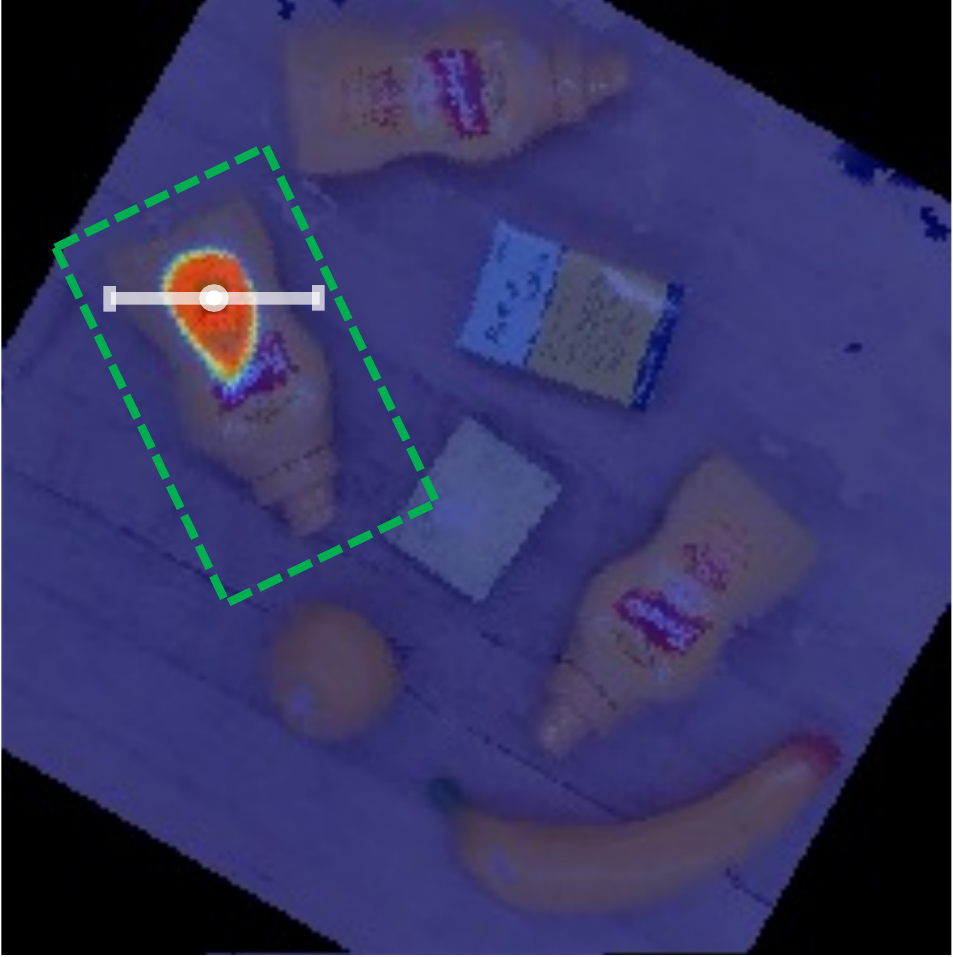}}\hfill
        {\includegraphics[width=0.24\textwidth]{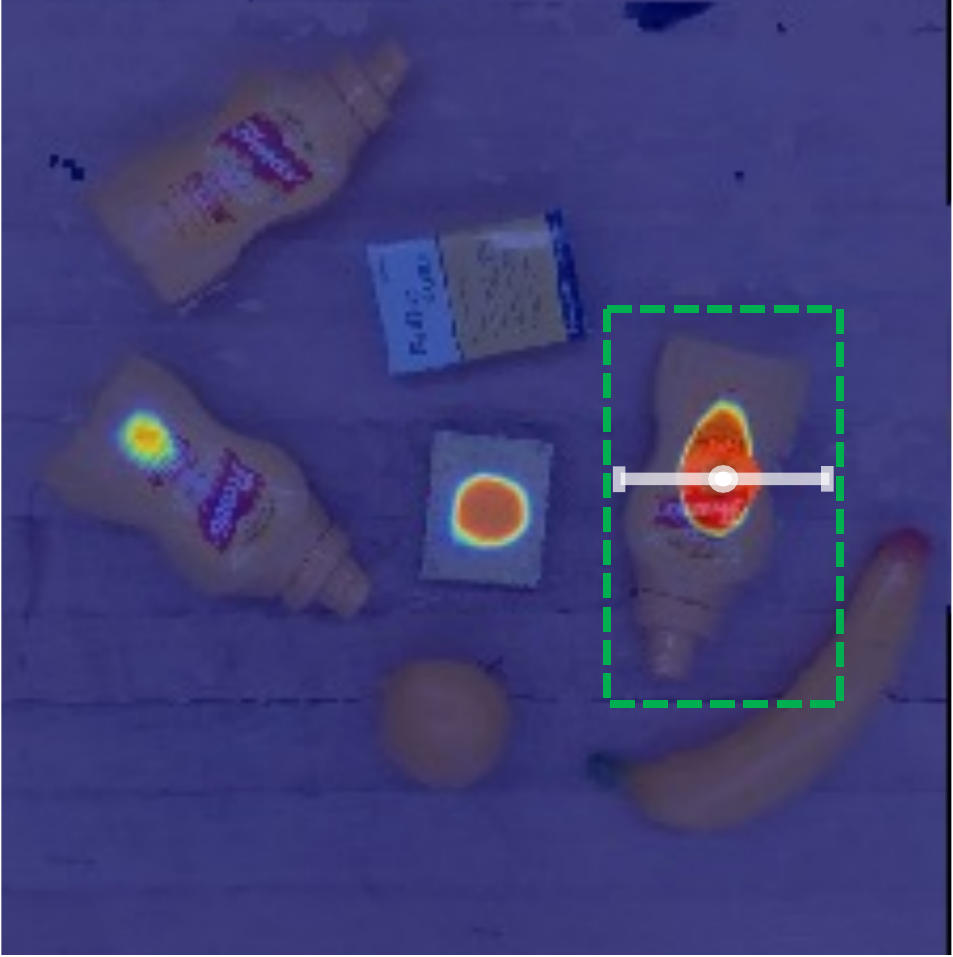}}\hfill
        {\includegraphics[width=0.24\textwidth]{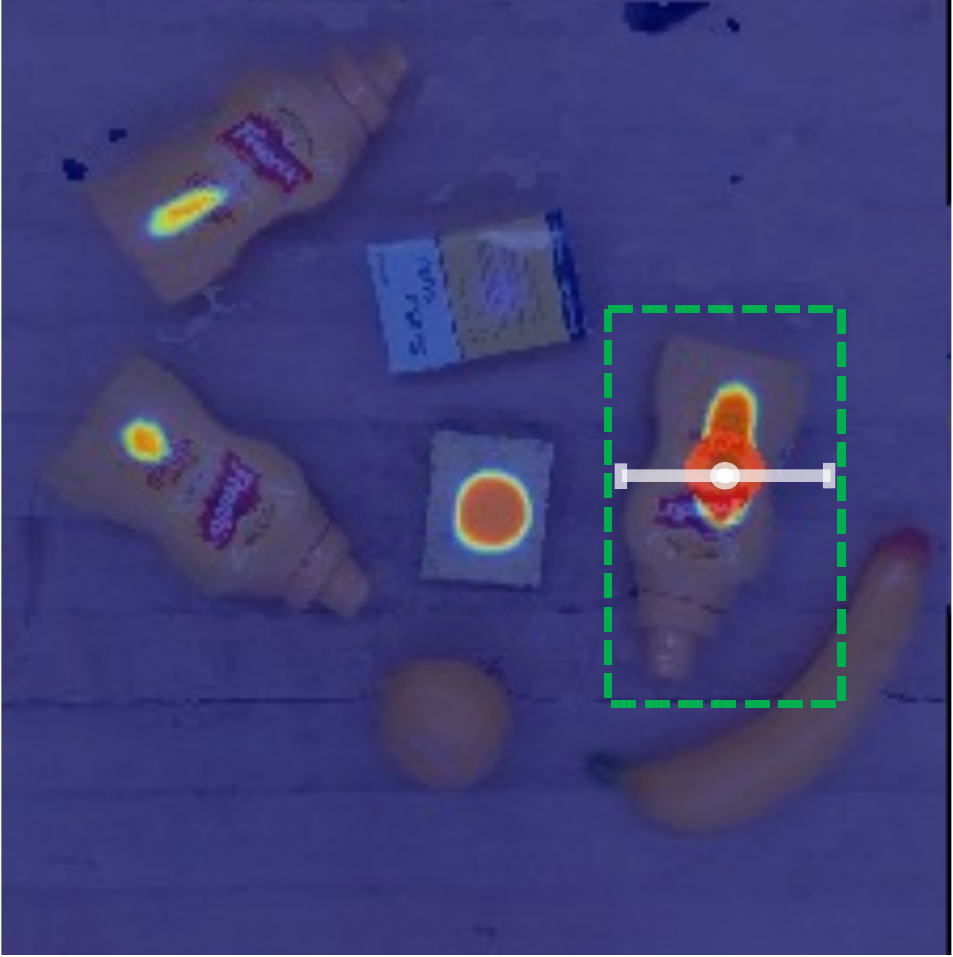}}%
        \vspace{-3pt}
        \caption{\textit{ETRG} grasping affordances}
        \label{fig:ETRG_s}
        \vspace{5pt}
    \end{subfigure}
    \begin{subfigure}{0.48\textwidth}
        {\includegraphics[width=0.24\textwidth]{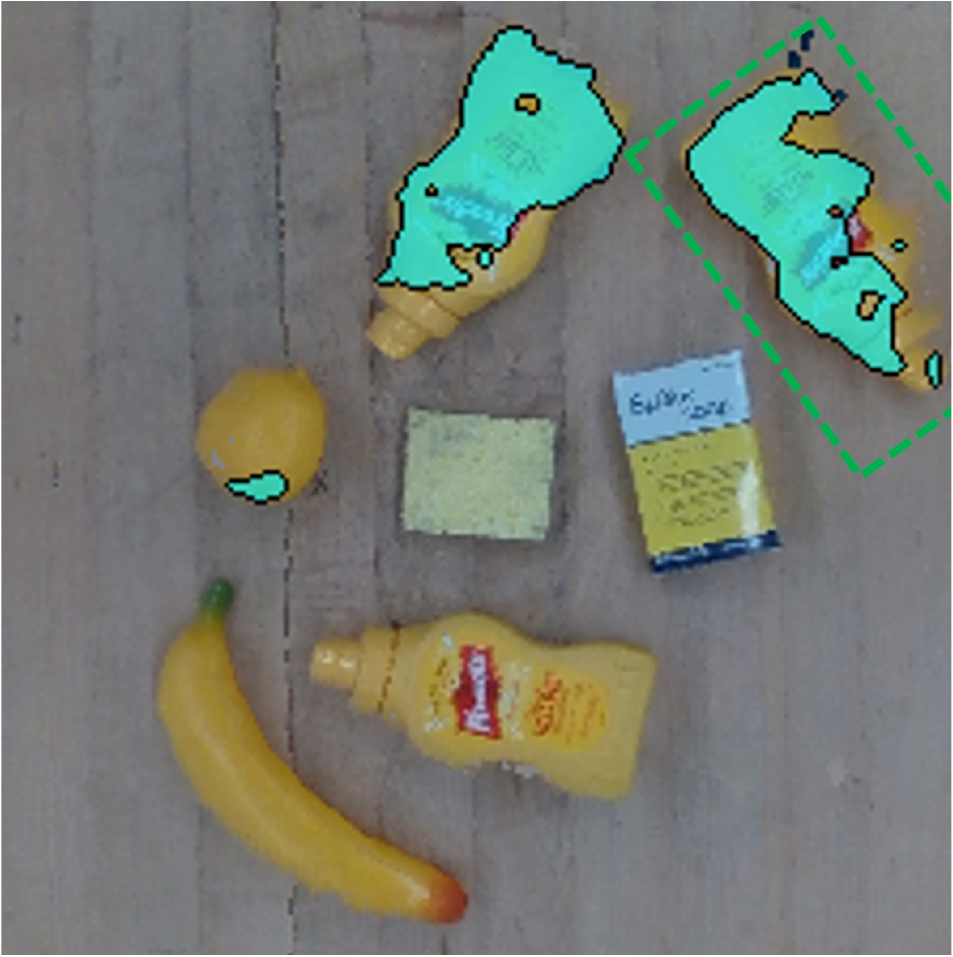}}\hfill
        {\includegraphics[width=0.24\textwidth]{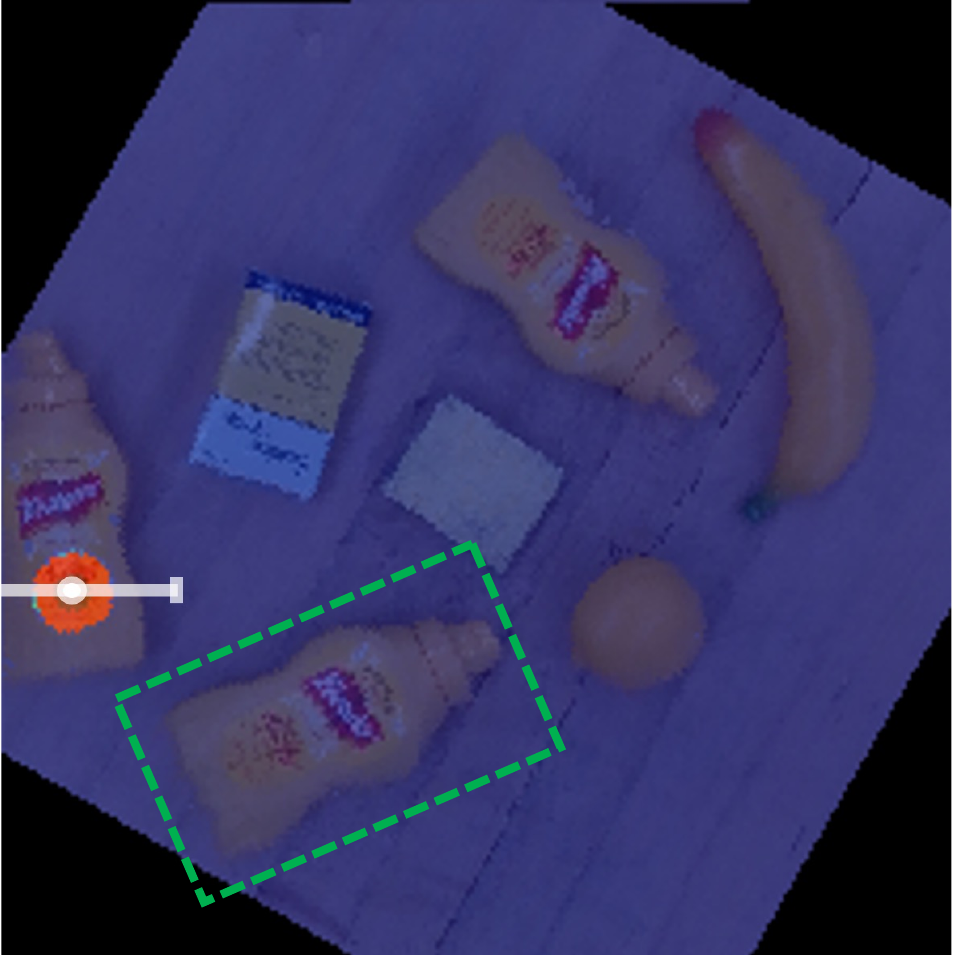}}\hfill
        {\includegraphics[width=0.24\textwidth]{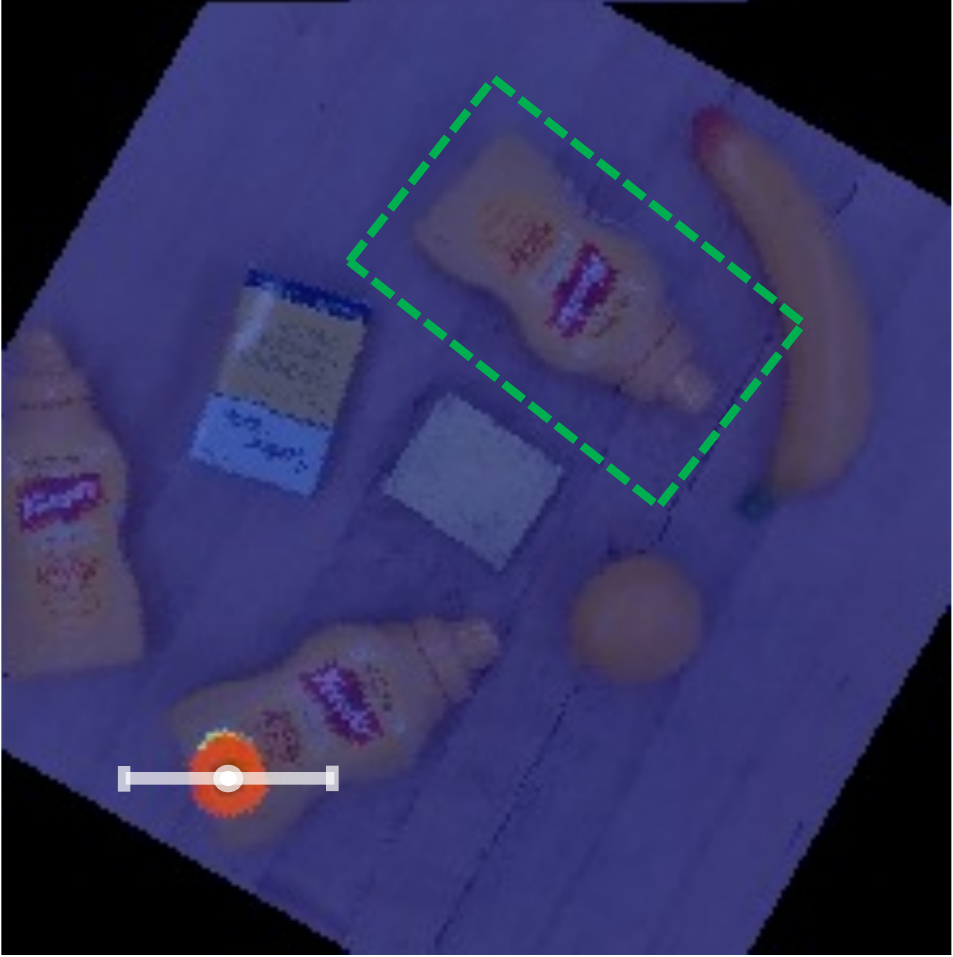}}\hfill
        {\includegraphics[width=0.24\textwidth]{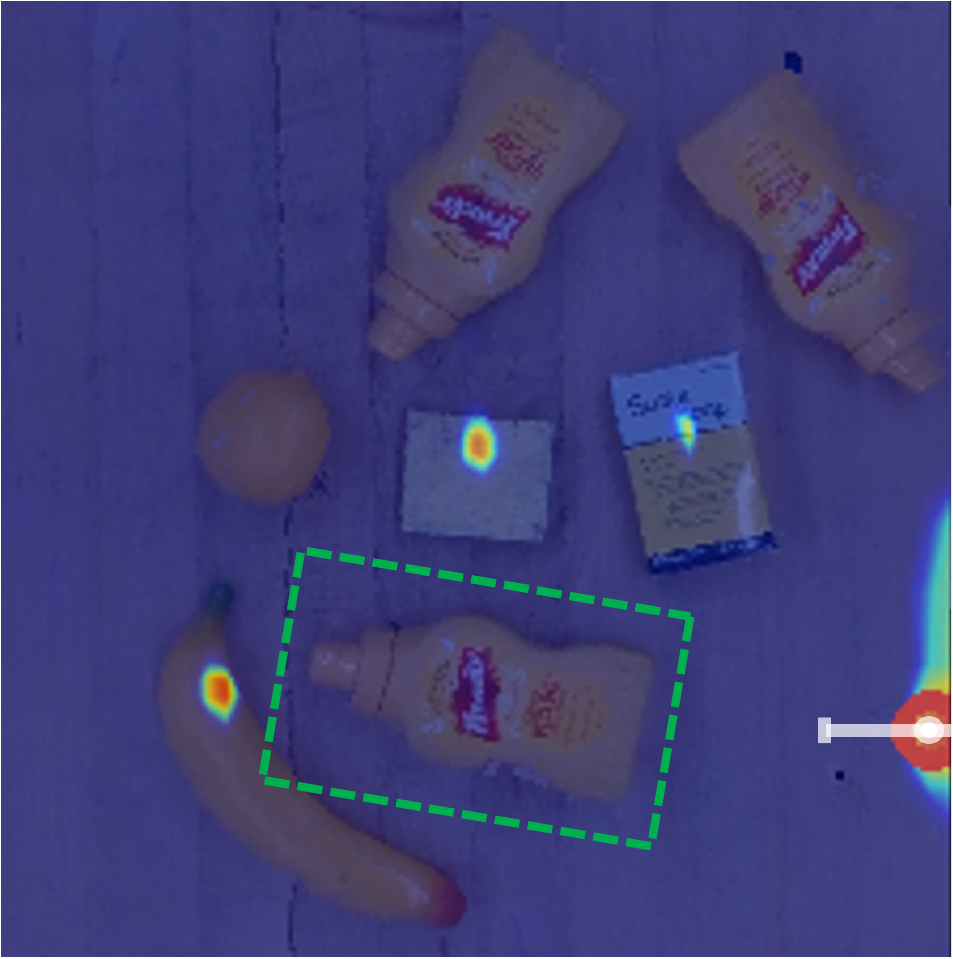}}%
        \vspace{-3pt}
        \caption{\textit{OGRG-{nodepth}} grounding and grasping maps}
        \label{fig:OGRG_nd_s}
        \vspace{5pt}
    \end{subfigure}
    \begin{subfigure}{0.48\textwidth}
        {\includegraphics[width=0.24\textwidth]{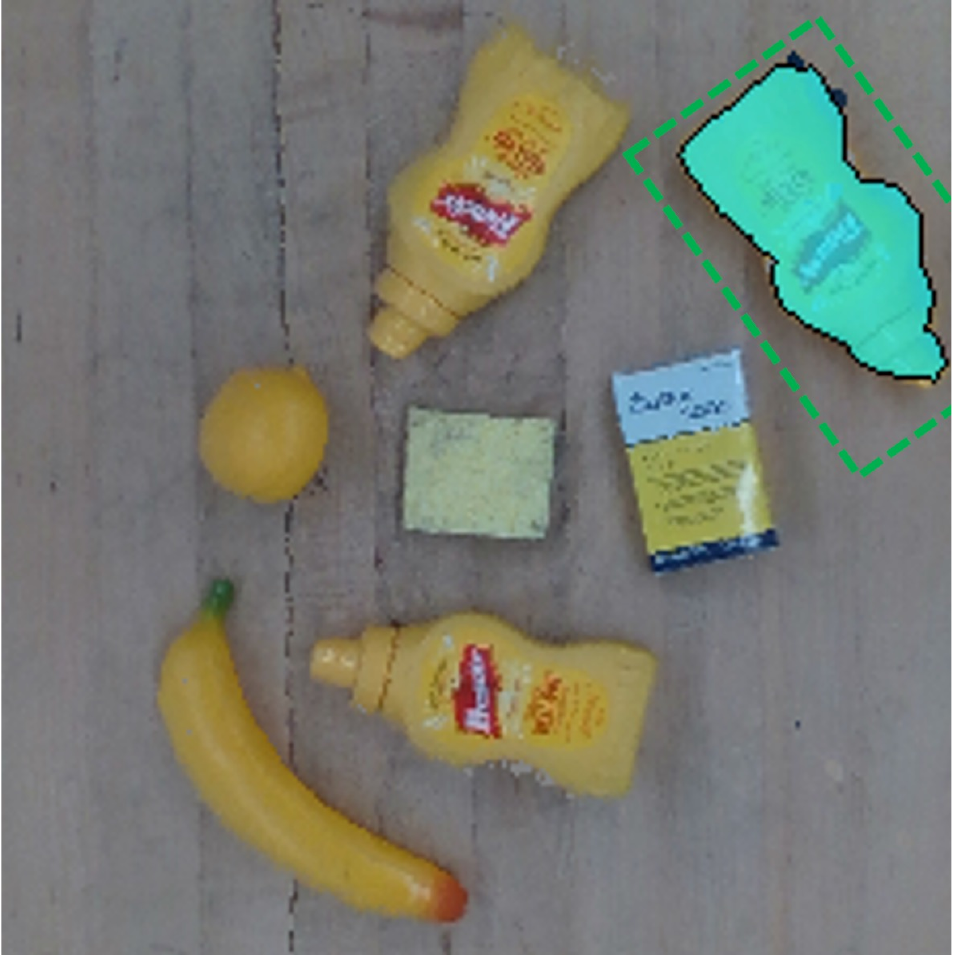}}\hfill
        {\includegraphics[width=0.24\textwidth]{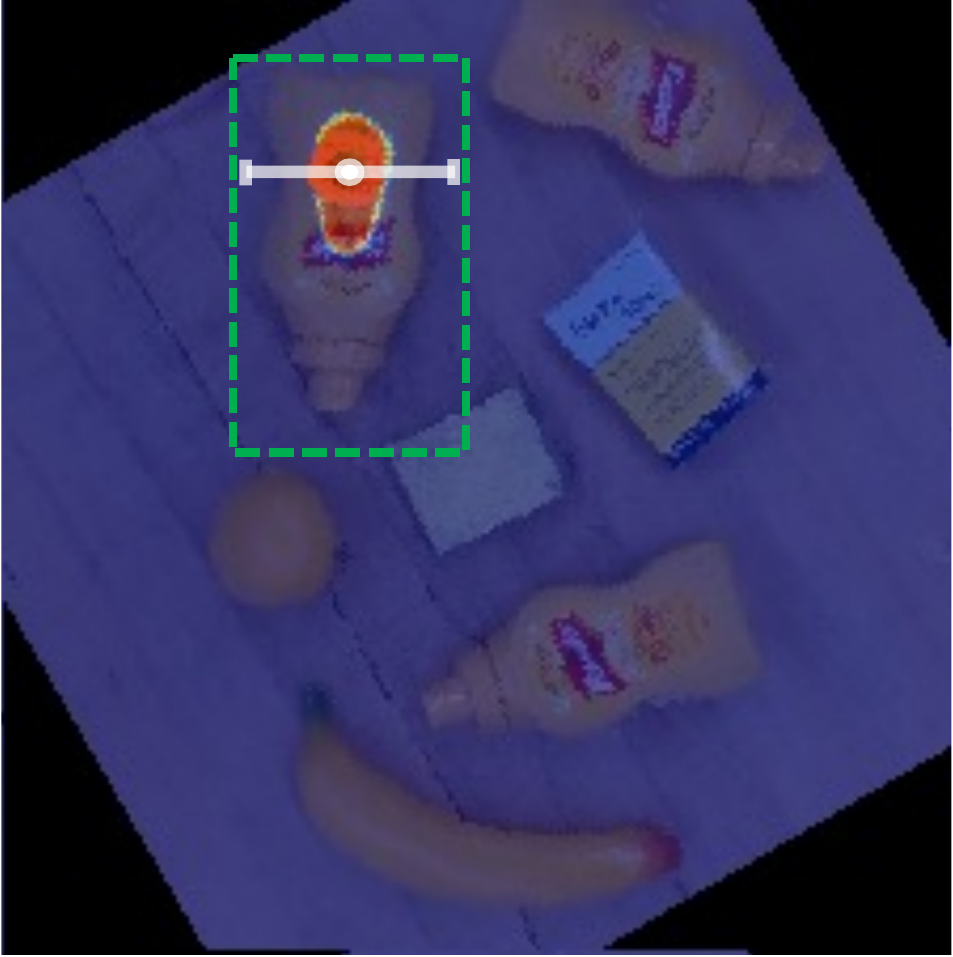}}\hfill
        {\includegraphics[width=0.24\textwidth]{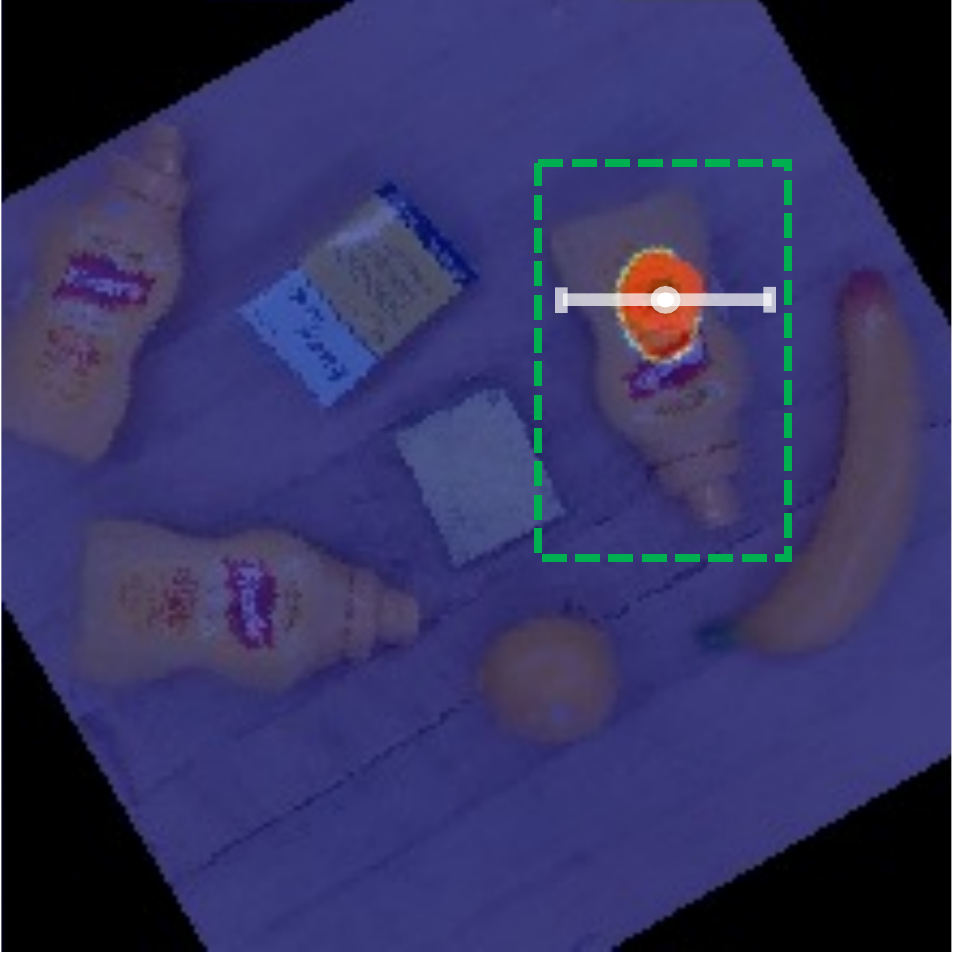}}\hfill
        {\includegraphics[width=0.24\textwidth]{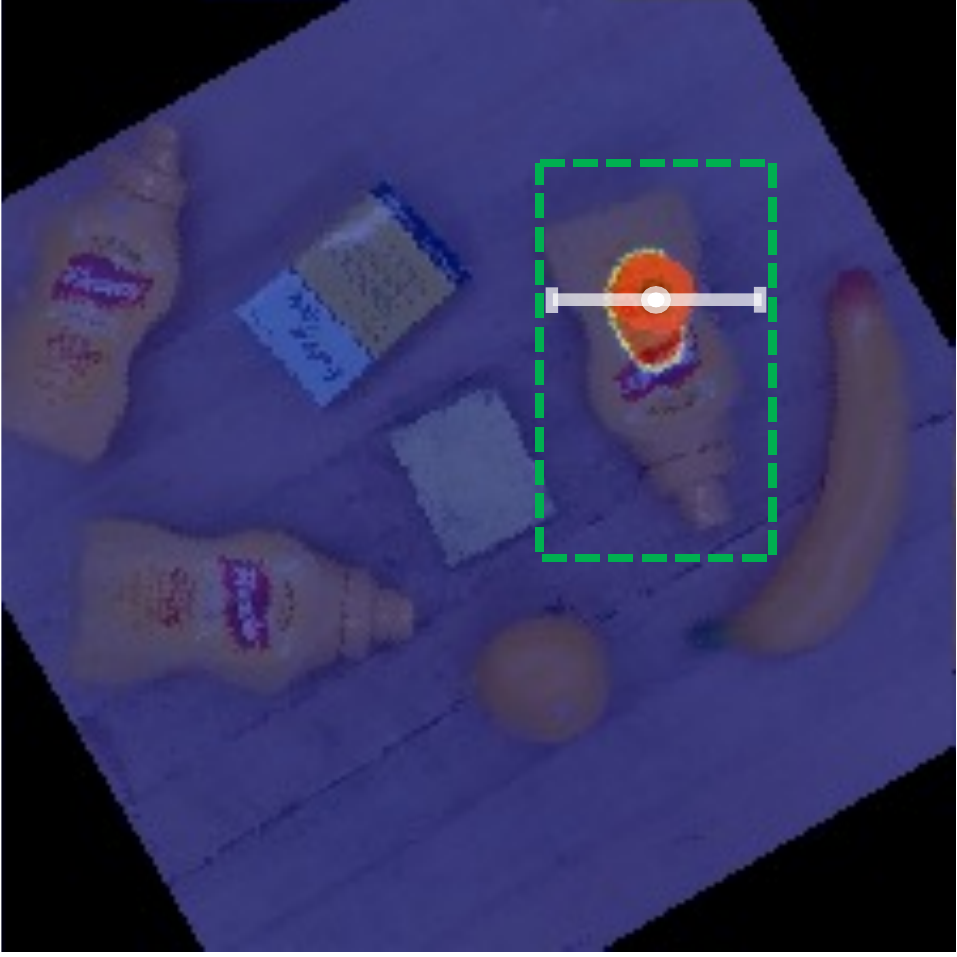}}%
        \vspace{-3pt}
        \caption{\textit{OGRG \textbf{(Ours)}} grounding and grasping maps}
        \label{fig:OGRG_s}
        \vspace{2pt}
    \end{subfigure}
    \caption{\textbf{Visualization of grounding masks and grasp affordance for challenging real robot spatial reasoning.} The green bounding boxes highlight the correct language-referred target object. The visualization in the first column shows the original scene arrangement and the grounding masks, while the rest of the three columns indicate the grasp affordances corresponding to spatial language descriptions on the same scene. The language expressions from left to right are: (1) grasp the upper right mustard bottle; (2) the mustard bottle that is to the top center of the workspace; (3) pass the mustard bottle that is to the lower right of the sponge; and (4) pass the mustard bottle below the sponge. Our proposed method demonstrates strong grounding capability, rapid adaptation, and accurate grasp pose predictions.}
    \label{fig:real-s}
\end{figure}

Fig. \ref{fig:real-ns} illustrates the qualitative results on general object attribute reasoning with distinct objects. Compared to the state-of-the-art Attribute-Grasp method \cite{yang2024attribute}, the proposed OGRG-RGA approach accurately localizes the target and generates precise grasp poses with distinct rotation angles (Fig. \ref{fig:OGRG_ns}). In contrast, Attribute-Grasp (Fig. \ref{fig:attrg_ns}) struggles when object candidates have similar colors or shapes as the target. Fig. \ref{fig:real-s} presents visualizations from challenging real robot spatial reasoning experiments. The OGRG-RGA method (Fig. \ref{fig:OGRG_s}) produces clean and correctly focused grounding and grasping affordance maps. While ETRG (Fig. \ref{fig:ETRG_s}) successfully localizes language-referred objects, it predicts high affordance values on incorrect objects. Additionally, OGRG-nodepth fails in both grounding and grasping tasks under these conditions.

Across 24 grasp attempts for each baseline, the qualitative evaluation of real robot grasping success rates is presented in Table \ref{real:grasp}. Due to the challenge of obtaining ground-truth target object masks in the real robot setup, a grounding accuracy metric is introduced to evaluate whether the grasping motion in each scene aligns with the target object. The proposed OGRG-RGA model achieves the highest grounding and grasping success rates, validating the effectiveness of its design, including the Bi-Aligner and depth fusion components. Despite the small dataset size used for fine-tuning, the model efficiently adapts to new scenes with novel objects.

\begin{table}[t]
\vspace{-1mm}
\caption{\textbf{OGRG-RGA real robot experiment results.} We evaluate the grounding accuracy and the grasping success rate under different scenes with challenging objects.}
\vspace{-1mm}
\label{real:grasp}
\begin{center}
\begin{tabular}{lcc}
\toprule
   Methods & Grounding (\%) & Grasp Succ. (\%) \\
\midrule
ETRG~\cite{yu2024parameter} & 75.0 & 62.5  \\
OGRG-{nodepth} & 75.0 & 33.3  \\
\textbf{OGRG (Ours)} & \textbf{87.5} & \textbf{70.8} \\
\bottomrule
\end{tabular}
\end{center}
\vspace{-6mm}
\end{table}

%% file: 05_conclusion.tex
\section{Conclusion}

We introduced OGRG, a framework that aligns visual and language features through a bidirectional aligner without relying on pre-aligned vision–language models. By incorporating depth fusion, OGRG supports open-form, attribute-based grounding and demonstrates rapid adaptability to new scenes and novel objects. On both RGS and RGA benchmarks, it achieves competitive performance in language-guided grounding and grasping. Although our experiments focus on planar grasps using a parallel-jaw gripper, the proposed approach is embodiment-agnostic and can be transferred to humanoid tabletop manipulation tasks. Furthermore, OGRG facilitates cross-embodiment grasping data collection by reusing language-grounded supervision across different end-effectors.

While OGRG achieves strong results on RGS and RGA tasks, our current evaluation is limited to common household objects with regular geometries in planar grasping settings. The experiments assume a fixed camera viewpoint, a parallel-jaw gripper, and relatively uncluttered tabletop environments. Addressing scenarios involving complex object geometries, severe occlusions, and diverse gripper configurations remains an important direction for future work.